\documentclass[11pt,a4paper]{article}
\usepackage[utf8]{inputenc}
\usepackage[T1]{fontenc}
\usepackage{mathptmx,textcomp}
\usepackage{microtype}
\usepackage[margin=2.4cm]{geometry}
\usepackage{amsmath,amssymb}
\usepackage{graphicx}
\usepackage{booktabs,longtable,array}
\usepackage[font=small,labelformat=empty,justification=justified,singlelinecheck=false]{caption}
\usepackage{enumitem}
\usepackage{float}
\usepackage[hidelinks]{hyperref}
\usepackage{setspace}
\setlist{itemsep=1pt,topsep=3pt}
\begin{document}
\begin{center}
{\LARGE\bfseries A richness-sensitive Acoustic Biodiversity Index from species-level detections: classifier benchmarking, false-positive control and validation against standard soundscape indices in an African savanna and an urban zoo\par}
\vspace{1em}

Muhammad Saeed\textsuperscript{a,*}, Dmitry Mikhaylov\textsuperscript{a}, Mohamed H. Alhosani\textsuperscript{a}, Yasser~F.~Al~Wahedi\textsuperscript{a}, Rory Pilossof\textsuperscript{b}, Ahmed Arabi\textsuperscript{c}, Walid Shaaba\textsuperscript{c}\\
\textsuperscript{a} Abu Dhabi Maritime Academy, P.O. Box 54477, Abu Dhabi, United Arab Emirates\\
\textsuperscript{b} My Trees Trust, 88 Broadlands Road, Mount Pleasant, Harare, Zimbabwe\\
\textsuperscript{c} Emirates Park Zoo and Resort, Abu Dhabi, United Arab Emirates\\
\textsuperscript{*} Corresponding author: muhammad.saeed@adports.ae (M. Saeed)\\
\end{center}

\section*{Abstract}
Acoustic indices are widely used as rapid proxies for biodiversity, but they summarise the distribution of sound energy rather than the species producing it and often track richness poorly. We revise and test an Acoustic Biodiversity Index (ABI) built from species-level detections. Its primary form, ABI\textsubscript{E}, is Shannon diversity of detections normalised by the regional species pool expected at the site and season (H / ln S\textsubscript{ref}); temporal stability T is reported as a separate component. We first benchmarked four classifiers on a strongly labelled tropical dataset (1,216 calls, 24 classes) with recording-grouped cross-validation. BirdNET embeddings + logistic regression performed best (macro-F1 0.750 $\pm$ 0.029), and random frame-level splits inflated accuracy from 0.41 to 0.70. We then applied BirdNET v2.4 with a location and season filter to recordings from a riverine savanna in the Zambezi Valley, Zimbabwe, and from three habitats at Emirates Park Zoo and Resort, Abu Dhabi, United Arab Emirates. A stricter post-hoc location filter removed 517 high-confidence detections of 9 implausible species, all at the zoo. After diel-matched rarefaction to equal effort (6 h), the savanna site held 21 species (\textsuperscript{1}D = 9.83; ABI\textsubscript{E} = 0.38) against 1.3–13 species (ABI\textsubscript{E} = 0.03–0.21) at the zoo sites. Across 149 recorded hours, ABI\textsubscript{E} correlated strongly with richness (Spearman $\rho$ = 0.93), whereas six standard indices did not (best: ACI, $\rho$ = 0.48). The stability term was lower at the savanna site (0.39) than at the zoo (0.65–0.67), showing that it penalises the natural diel cycle of bird activity. Species-level indices with explicit false-positive control, a declared regional reference pool and effort standardisation provide a more interpretable biodiversity indicator than energy-based acoustic indices.

\textbf{Keywords:} passive acoustic monitoring; acoustic indices; BirdNET; Hill numbers; soundscape ecology; false positives; Zambezi Valley

\subsection*{Highlights}
\begin{itemize}
\item Species-level detections with a regional reference pool track richness ($\rho$ = 0.93)
\item Six standard acoustic indices tracked hourly richness weakly (|$\rho$| $\leq$ 0.48)
\item A location filter removed 517 high-confidence false detections at a zoo
\item At equal effort a savanna site held 21 bird species vs 1.3–13 at zoo sites
\item Temporal-stability terms penalise the natural diel cycle and should stand apart
\end{itemize}
\clearpage
\section*{1. Introduction}
Passive acoustic monitoring (PAM) records whole soundscapes continuously, at low cost and with little disturbance, and has become a core tool for tracking biodiversity in places that are hard to survey on foot (Gibb et al., 2019; Sugai et al., 2019). Autonomous recorders now routinely produce thousands of hours of audio per deployment. At that volume, manual listening is impossible and ecological inference depends on automated summaries of the recordings.

The most common summaries are acoustic indices. The Acoustic Complexity Index (ACI; Pieretti et al., 2011) measures intensity changes between adjacent time frames; the Acoustic Diversity and Evenness Indices (ADI, AEI; Villanueva-Rivera et al., 2011) measure how evenly energy is spread across frequency bands; the Bioacoustic Index (BI; Boelman et al., 2007) measures energy in the 2–8 kHz band typical of bird song; the Normalised Difference Soundscape Index (NDSI; Kasten et al., 2012) contrasts biophony with anthrophony; and acoustic entropy (H; Sueur et al., 2008) combines temporal and spectral entropy. These indices are fast, need no training data and can be computed for any recording.

However, acoustic indices describe how sound energy is distributed in time and frequency, not which species are calling. A meta-analysis of more than 1,000 index–biodiversity relationships found moderate and highly variable correlations (Alcocer et al., 2022). Indices respond strongly to rain, wind, insects and anthropogenic noise, and their behaviour differs between ecosystems (Fairbrass et al., 2017; Buxton et al., 2018; Sethi et al., 2020). A high index value therefore does not necessarily indicate high diversity.

Deep-learning classifiers now offer a second route: identifying species directly. BirdNET recognises more than 6,000 bird species and a set of non-biological sound classes, and includes a location and season meta-model that restricts predictions to species expected at a given site and week (Kahl et al., 2021). Its internal embeddings also transfer well to new classification tasks with little training data (Ghani et al., 2023). Species-level detections can be turned directly into community metrics such as richness and Hill numbers (Hill, 1973; Jost, 2006; Chao et al., 2014), which ecologists already know how to interpret.

Two problems limit this route in practice. First, classifiers produce false positives, sometimes with high confidence, and these inflate richness unless detections are filtered and verified (Wood and Kahl, 2024). Second, classifier performance is often overestimated because evaluation splits place short frames from the same recording, or even the same call, in both training and test data (Roberts et al., 2017; Stowell, 2022). An indicator built on detections inherits both problems.

A detection-based index also has to be defined carefully. Normalising Shannon diversity by the \textit{observed} richness gives Pielou's evenness J$'$ (Pielou, 1966), which is insensitive to richness. Combining it with an unweighted temporal coefficient of variation makes the index dominated by rare species, which have the largest CV, and dependent on the number of time bins. In our earlier formulation of the ABI, these properties caused the index to \textit{decrease} as true richness increased (Section 3.7).

Here we revise the ABI and test every step of its production, from classifier evaluation to field application. Our aims were to: (i) benchmark species classifiers with leakage-free, recording-grouped evaluation and quantify the effect of leakage and domain shift; (ii) derive species-level detections with explicit false-positive control in two contrasting soundscapes, a riverine savanna in the Zambezi Valley (Zimbabwe) and a zoo in Abu Dhabi (United Arab Emirates); (iii) define a bounded, richness-sensitive ABI and characterise its behaviour by simulation; (iv) test its robustness to sampling effort, confidence threshold, time-bin width and filter strictness; and (v) compare it with six standard acoustic indices at hourly resolution. The two study areas were chosen as a proof of concept across highly contrasting environments, not as an ecological comparison of two regions: a useful indicator must at least separate a natural savanna from a managed urban setting, and the extreme contrast lets the behaviour of each index be examined without ambiguity about which site is richer.

\section*{2. Materials and methods}
The two study areas are mapped in Fig. 1 and described in Table 1; Fig. 2 summarises the workflow. The analysis has three steps (classifier benchmark, field soundscapes, ABI analysis), implemented in a Python package described in Section 2.8.

\begin{figure}[!htbp]\centering
\includegraphics[width=\textwidth,height=0.78\textheight,keepaspectratio]{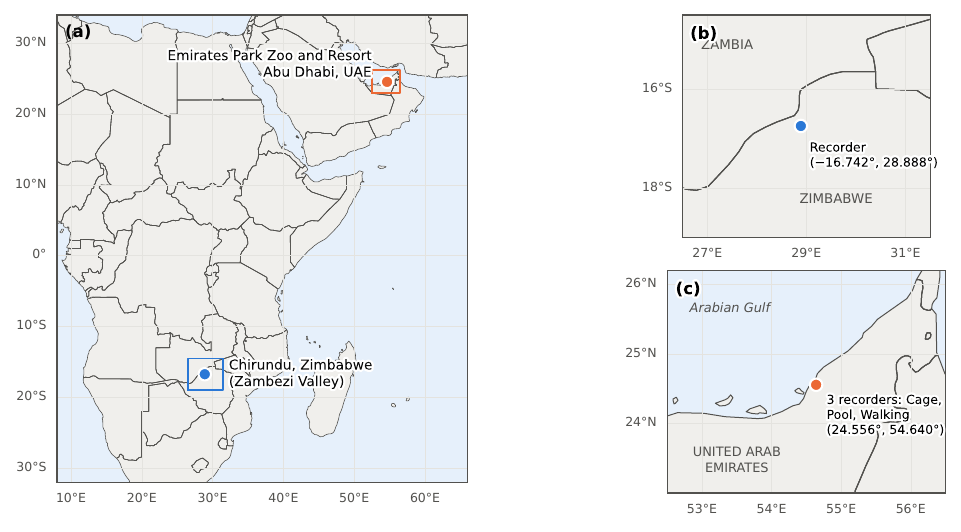}\par\medskip
\caption{\textbf{Fig. 1.} Location of the two study areas: Chirundu in the Zambezi Valley, northern Zimbabwe (riverine savanna; one recorder), and Emirates Park Zoo and Resort, Abu Dhabi, United Arab Emirates (three recorders in a cage enclosure, at the hippopotamus pool and along a visitor walkway).}
\end{figure}

\begin{figure}[!htbp]\centering
\includegraphics[width=\textwidth,height=0.78\textheight,keepaspectratio]{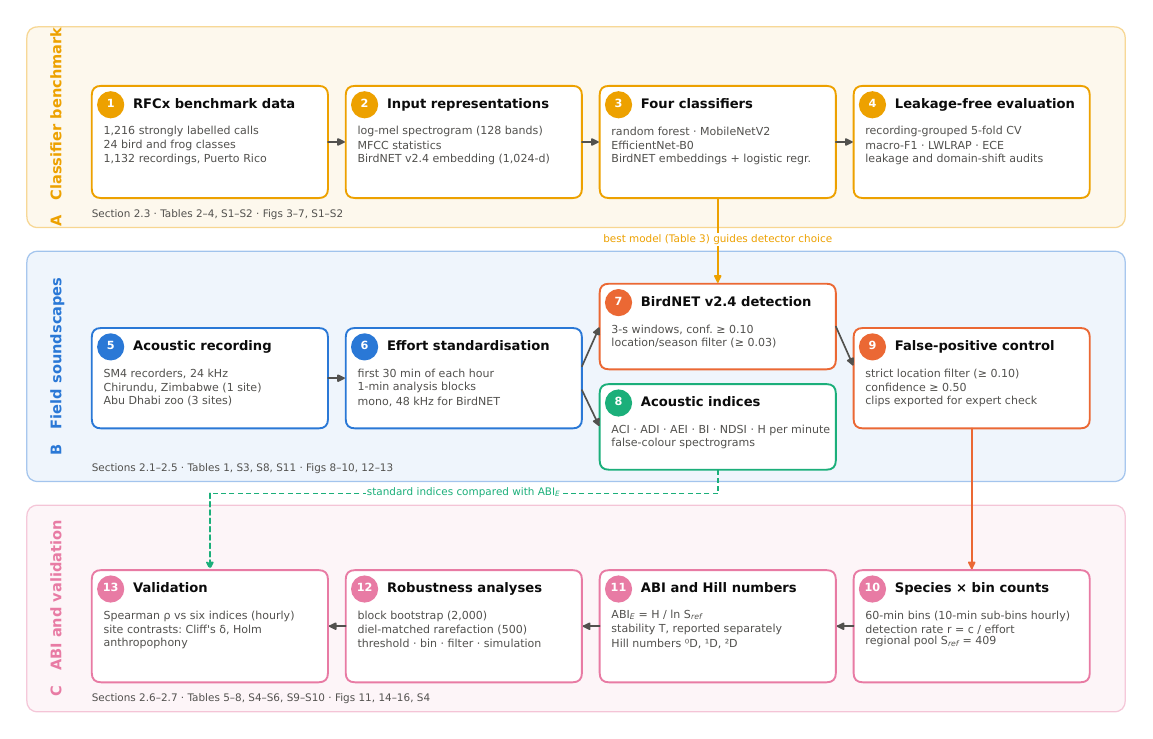}\par\medskip
\caption{\textbf{Fig. 2.} Analysis workflow. Field recordings are standardised to 30 min h\textsuperscript{$-$}\textsuperscript{1} and analysed in 1-min blocks for six acoustic indices and for BirdNET species detections with a location and season filter. A separate classifier benchmark on strongly labelled data evaluates four models with recording-grouped cross-validation. Detections feed the ABI, Hill numbers and their uncertainty, sensitivity and simulation analyses, which are then compared with the acoustic indices.}
\end{figure}

\subsection*{2.1 Study sites}
\textbf{Chirundu, Zimbabwe.} The savanna site lies in the middle Zambezi Valley near Chirundu ($-$16.742\textdegree{}, 28.888\textdegree{}), in woodland managed by My Trees Trust. The vegetation is riverine woodland and mopane savanna close to the Zambezi River. My Trees Trust runs restoration, anti-poaching and conservation programmes in the middle Zambezi Valley, and its conservation areas are managed under a Forest Stewardship Council plan (My Trees Trust, 2023). Recordings were made on 1–2 February 2024, during the wet season, when many resident and intra-African migrant birds are breeding.

\textbf{Emirates Park Zoo and Resort, Abu Dhabi, UAE.} The contrasting site is a zoological park in Abu Dhabi (24.556\textdegree{}, 54.640\textdegree{}) in the hyper-arid lowlands of the Arabian Peninsula. Three recorders were placed in different habitats: a caged animal enclosure (Cage), the hippopotamus pool exhibit (Pool) and a visitor walkway with ornamental planting (Walking). Recordings were made on 27–29 July 2025, in the hot season, when wild bird activity in the region is low and resident urban species dominate. The zoo soundscape also includes captive animals, visitors, maintenance machinery and water pumps. Recordings were made with the permission of the zoo management.

{\small\setlength{\tabcolsep}{4pt}
\begin{longtable}{p{2.1cm}p{2.3cm}p{3.1cm}p{2.0cm}p{2.0cm}cc}
\caption{\textbf{Table 1.} Study sites, recorders and analysed effort. All recorders sampled at the rate given in Table S3; effort is the analysed duration after standardisation to 30 min h\textsuperscript{$-$}\textsuperscript{1}.}\\
\toprule
\textbf{Site} & \textbf{Coordinates} & \textbf{Habitat} & \textbf{Recorder} & \textbf{Dates} & \textbf{Files} & \textbf{Analysed (h)} \\
\midrule
\endfirsthead
\toprule
\textbf{Site} & \textbf{Coordinates} & \textbf{Habitat} & \textbf{Recorder} & \textbf{Dates} & \textbf{Files} & \textbf{Analysed (h)} \\
\midrule
\endhead
\bottomrule
\endlastfoot
Chirundu (ZW) & $-$16.742, 28.888 & Riverine savanna, Zambezi Valley & SM4 & 1–2 Feb 2024 & 12 & 6.0 \\
Zoo – Cage & 24.556, 54.640 & Caged animal enclosure & SM4 S4A27792 & 27–29 Jul 2025 & 27 & 13.1 \\
Zoo – Pool & 24.556, 54.640 & Hippopotamus pool exhibit & SM4 S4A27903 & 27–29 Jul 2025 & 54 & 26.8 \\
Zoo – Walking & 24.556, 54.640 & Visitor walkway & SM4 S4A27910 & 27–29 Jul 2025 & 55 & 27.0 \\
\end{longtable}
}

\subsection*{2.2 Acoustic recording and sampling effort}
All sites were recorded with Wildlife Acoustics Song Meter SM4 units with built-in omnidirectional microphones. All files were sampled at 24 kHz (Table S3), so the usable bandwidth is limited to 12 kHz. The zoo recorders were stereo and their channels were averaged. The Chirundu recorder made files at 15:11 and 18:11 on 1 February and hourly from 02:11 to 11:11 on 2 February (12 files, 12.0 h). The zoo recorders ran hourly for 1–2.3 days. The recorders followed different schedules, so effort was standardised: the first 30 min of every hourly file was analysed, matching the 30-min duty cycle of the Chirundu recorder. Audio was processed in 1-min blocks, mixed to mono and resampled to 48 kHz where a model required it. The analysed effort was 6.0 h at Chirundu, 13.1 h at Cage, 26.8 h at Pool and 27.0 h at Walking.

\subsection*{2.3 Classifier benchmark}
\subsubsection*{2.3.1 Benchmark data}
Field recordings from Chirundu are not annotated, so classifiers were compared on a public, strongly labelled dataset: the true-positive annotations of the Rainforest Connection (RFCx) Species Audio Detection challenge (Kaggle, 2021). It contains 1,216 annotated calls of 24 bird and frog classes in 1,132 one-minute recordings from Puerto Rico, each with start and end times and frequency limits (Table 2; Fig. 3).

\subsubsection*{2.3.2 Features and models}
Three input representations were computed for each annotated call:

\begin{itemize}
\item \textbf{Log-mel spectrogram:} 128 mel bands between 50 Hz and 16 kHz (FFT size 2,048, hop 1,024 samples at 48 kHz) of a 12-s window centred on the call. The networks were trained on random 10-s crops and tested on the central 10 s.
\item \textbf{MFCC statistics:} the mean and standard deviation of 20 mel-frequency cepstral coefficients and their deltas, plus spectral centroid, bandwidth and flatness, computed over the annotated call.
\item \textbf{BirdNET embedding:} the 1,024-dimensional embedding of BirdNET v2.4 for the 3-s window centred on the call.
\end{itemize}
Four models were compared: (1) a random forest (500 trees, balanced class weights) on MFCC statistics; (2) MobileNetV2 (Sandler et al., 2018) and (3) EfficientNet-B0 (Tan and Le, 2019), both initialised from ImageNet weights and fine-tuned on log-mel spectrograms; and (4) logistic regression (L2, C = 0.5, balanced class weights) on standardised BirdNET embeddings. The networks were trained with AdamW (weight decay 0.01), a one-cycle learning-rate schedule (maximum 10\textsuperscript{$-$}\textsuperscript{3}), label smoothing of 0.1, mixed precision and SpecAugment (two frequency and two time masks; Park et al., 2019) for up to 40 epochs. Training stopped early when the macro-F1 on an inner validation split (15\% of training recordings, grouped by recording) had not improved for 10 epochs, and the best checkpoint was kept.

\subsubsection*{2.3.3 Evaluation}
All models were evaluated with stratified group 5-fold cross-validation grouped by recording, so no recording contributed to both training and test data in any fold. We report accuracy, balanced accuracy, macro-F1, top-3 accuracy, label-weighted label-ranking average precision (LWLRAP, equivalent here to the mean reciprocal rank of the true class) and expected calibration error (ECE), each as the mean and standard deviation over folds with a 95\% confidence interval. Per-class F1 values were computed from the pooled out-of-fold predictions. For K classes with per-class precision P\textsubscript{k} and recall R\textsubscript{k},

\begin{equation}\mathrm{F1}_{\mathrm{macro}} = \frac{1}{K}\sum_{k=1}^{K}\frac{2\,P_k R_k}{P_k + R_k}\tag{1}\end{equation}
LWLRAP reduces, for single-label data, to the mean reciprocal rank of the true class over the N test calls,

\begin{equation}\mathrm{LWLRAP} = \frac{1}{N}\sum_{j=1}^{N}\frac{1}{\mathrm{rank}_j}\tag{2}\end{equation}
and ECE is the weighted gap between confidence and accuracy over M = 15 equal-width confidence bins B\textsubscript{m},

\begin{equation}\mathrm{ECE} = \sum_{m=1}^{M}\frac{|B_m|}{N}\,\bigl|\mathrm{acc}(B_m) - \mathrm{conf}(B_m)\bigr|\tag{3}\end{equation}
\subsubsection*{2.3.4 Leakage and domain-shift audits}
Two audits tested how evaluation choices and domain shift affect apparent performance. The leakage audit repeated a frame-level random-forest analysis (13 MFCCs per 512-sample frame, 200 trees) with two protocols: a random 80/20 split of frames, as in common practice, and a recording-grouped 5-fold split. The domain-shift audit applied the five fold models of the best CNN, as an ensemble, to 10-s windows of the Chirundu recordings and compared the distribution of their maximum softmax probability with that of out-of-fold RFCx predictions. Because none of the 24 RFCx classes occurs in the Zambezi Valley, a well-behaved classifier should be uncertain on these windows.

\subsection*{2.4 Species detection in the field recordings}
\subsubsection*{2.4.1 BirdNET detection}
Field recordings were analysed with BirdNET v2.4 (TFLite model; Kahl et al., 2021) in consecutive 3-s windows. All predictions with confidence $\geq$ 0.10 were stored. During detection, predictions were restricted to species with a meta-model score $\geq$ 0.03 for each site's coordinates and week of the year (BirdNET's default). This gave 455 candidate species for Chirundu in early February. Eleven non-biological classes (dog, engine, environmental, fireworks, gun, human non-vocal, human vocal, human whistle, noise, power tools, siren) were always retained but kept separate from the community data. Their detection rate per analysed minute was used as a measure of anthropophony.

\subsubsection*{2.4.2 Stricter post-hoc location filter}
The default meta-model threshold is permissive by design. For each site we therefore queried the meta-model at a grid of thresholds (0.01–0.30) and recorded, for every species, the highest threshold at which it was still expected (its location score). Detections of species with a location score < 0.10 were removed before analysis. The regional reference pool S\textsubscript{ref} = 409 is the largest site species list at this threshold (Chirundu). The effect of the filter threshold was tested in a sensitivity analysis (Section 2.7.3).

\subsubsection*{2.4.3 Confidence threshold and expert verification}
Biological detections were retained at confidence $\geq$ 0.50. To allow expert verification, a stratified random sample of detections across species and confidence levels was exported as audio clips with spectrograms and a verification template. When verified clips are available, the workflow fits a logistic model of the probability that a detection with confidence c is correct, pooled and per species ($\geq$ 8 clips with both outcomes),

\begin{equation}P(\text{correct} \mid c) = \frac{1}{1 + \exp\!\left[-\left(\beta_0 + \beta_1\,\mathrm{logit}(c)\right)\right]}\tag{4}\end{equation}
and derives the threshold c* that gives a target precision $\pi$ = 0.90 (Wood and Kahl, 2024),

\begin{equation}c^{*} = \mathrm{logit}^{-1}\!\left(\frac{\mathrm{logit}(\pi) - \beta_0}{\beta_1}\right)\tag{5}\end{equation}
These species-specific thresholds then replace the global value.

\subsection*{2.5 Acoustic indices}
For every analysed minute we computed six standard indices with scikit-maad (Ulloa et al., 2021): ACI, ADI, AEI, BI, NDSI and acoustic entropy H = H\textsubscript{t} $\times$ H\textsubscript{f}, the product of temporal and spectral entropy (Sueur et al., 2008). Spectrograms were limited to the Nyquist frequency of the recordings (12 kHz). ADI and AEI used 1-kHz bands from 0 to 12 kHz with a $-$50 dB threshold. BI used 2–8 kHz, the default of the original implementation. NDSI used 2–11 kHz for biophony and 1–2 kHz for anthrophony. We also computed the equivalent sound level (L\textsubscript{eq}, dBFS). Long-duration false-colour spectrograms (Towsey et al., 2014) were built by mapping per-band ACI, the fraction of time a band exceeded its median by 6 dB (event fraction) and mean spectral power to red, green and blue, each normalised by percentiles per site.

With S\textsubscript{f,t} the spectrogram amplitude in frequency bin f and time frame t, ACI sums, over frequency bins, the frame-to-frame amplitude change relative to the total amplitude in the bin (Pieretti et al., 2011):

\begin{equation}\mathrm{ACI} = \sum_{f}\frac{\sum_{t}\left|S_{f,t+1} - S_{f,t}\right|}{\sum_{t} S_{f,t}}\tag{6}\end{equation}
ADI and AEI are the Shannon entropy and the Gini coefficient of q\textsubscript{b}, the proportion of time–frequency cells above the $-$50 dB threshold in frequency band b, normalised over the B = 12 bands (Villanueva-Rivera et al., 2011):

\begin{equation}\mathrm{ADI} = -\sum_{b=1}^{B} q_b \ln q_b, \qquad \mathrm{AEI} = \frac{\sum_{i=1}^{B}\sum_{j=1}^{B}\left|q_i - q_j\right|}{2B\sum_{i=1}^{B} q_i}\tag{7}\end{equation}
BI is the area under the time-averaged spectrum in decibels above its minimum, between 2 and 8 kHz (Boelman et al., 2007), and NDSI contrasts the power $\beta$ in the biophony band with the power $\alpha$ in the anthrophony band (Kasten et al., 2012):

\begin{equation}\mathrm{BI} = \sum_{f=2\,\mathrm{kHz}}^{8\,\mathrm{kHz}}\left(\bar{L}_f - \min_{f}\bar{L}_f\right)\Delta f, \qquad \mathrm{NDSI} = \frac{\beta - \alpha}{\beta + \alpha}\tag{8}\end{equation}
Acoustic entropy is the product of the normalised entropies of the amplitude envelope A\textsubscript{t} (n\textsubscript{t} frames) and of the mean spectrum (n\textsubscript{f} bins), each scaled to sum to one (Sueur et al., 2008):

\begin{equation}H = H_t H_f, \qquad H_t = -\frac{\sum_{t} A_t \log_2 A_t}{\log_2 n_t}, \qquad H_f = -\frac{\sum_{f} \bar{S}_f \log_2 \bar{S}_f}{\log_2 n_f}\tag{9}\end{equation}
\subsection*{2.6 The Acoustic Biodiversity Index}
Let c\textsubscript{i,t} be the number of 3-s windows in which species i is detected in time bin t (t = 1, \ldots{}, n), and let e\textsubscript{t} be the analysed minutes in bin t. The share of all detections that belongs to species i is p\textsubscript{i} = $\Sigma$\textsubscript{t} c\textsubscript{i,t} / $\Sigma$\textsubscript{i,t} c\textsubscript{i,t}, S is the observed number of species and r\textsubscript{i,t} = c\textsubscript{i,t} / e\textsubscript{t} is the detection rate. Shannon diversity of the detections is

\begin{equation}H = -\sum_{i=1}^{S} p_i \ln p_i\tag{10}\end{equation}
The primary index normalises H by the logarithm of the regional reference pool rather than by the observed richness (Eq. 11):

\begin{equation}\mathrm{ABI}_E = \frac{H}{\ln S_{\mathrm{ref}}}\tag{11}\end{equation}
Because S\textsubscript{ref} is fixed, ABI\textsubscript{E} increases with richness as well as with evenness, and lies in [0, 1] (it reaches 1 only if every species in the pool is detected equally often). By contrast, Pielou's J$'$ = H / ln S is independent of richness. ABI\textsubscript{E} is intended as a standardised biodiversity indicator, which expresses the diversity of the detected community as a fraction of what the regional pool could support, rather than as a proxy for species richness alone; richness itself is reported alongside it as \textsuperscript{0}D.

Temporal stability is measured from the coefficient of variation of each species' detection rate across time bins,

\begin{equation}\mathrm{CV}_i = \frac{\mathrm{sd}_t\left(r_{i,t}\right)}{\mathrm{mean}_t\left(r_{i,t}\right)}\tag{12}\end{equation}
The largest possible value of CV\textsubscript{i}, reached when a species occurs in a single bin, is $\surd$(n $-$ 1). Dividing by this maximum makes the term independent of the number of bins, and weighting by p\textsubscript{i} prevents rare species from dominating:

\begin{equation}T = 1 - \sum_{i=1}^{S} p_i \, \frac{\mathrm{CV}_i}{\sqrt{n-1}}\tag{13}\end{equation}
T lies in [0, 1] and equals 1 when every species is detected at a constant rate. The combined index is

\begin{equation}\mathrm{ABI}_{\mathrm{rev}} = \mathrm{ABI}_E \times T\tag{14}\end{equation}
For comparison we also computed the earlier formulation,

\begin{equation}\mathrm{ABI}_{\mathrm{orig}} = \frac{H}{\ln S} \cdot \frac{1}{1 + \overline{\mathrm{CV}}}\tag{15}\end{equation}
where the mean CV is taken over species without weighting. Hill numbers of order q = 0, 1, 2 (Hill, 1973; Jost, 2006) were computed from the same counts,

\begin{equation}{}^{q}D = \left(\sum_{i=1}^{S} p_i^{\,q}\right)^{1/(1-q)}, \qquad {}^{1}D = \exp(H)\tag{16}\end{equation}
so that \textsuperscript{0}D is richness, \textsuperscript{1}D the effective number of common species and \textsuperscript{2}D the inverse Simpson index of dominant species.

\subsection*{2.7 Statistical analysis}
\subsubsection*{2.7.1 Uncertainty}
Site-level metrics used 60-min time bins. Confidence intervals (95\%) were obtained from 2,000 bootstrap resamples of whole time bins (a block bootstrap; Künsch, 1989), which preserves the dependence among detections within a bin. Observed richness can only decrease under resampling, so where the percentile interval excluded the point estimate it was extended to include it.

\subsubsection*{2.7.2 Effort standardisation}
Analysed effort differed up to 4.5-fold between sites. We therefore rarefied the zoo sites to the effort of the least-sampled site (Chirundu) with a diel-matched design: in each of 500 draws, time bins were sampled without replacement so that the hour-of-day composition matched Chirundu's exactly. Matching hours removes the confounding effect of the strong diel cycle in bird activity. We report the mean and the 2.5–97.5\% range across draws. Species accumulation curves were built from 100 random orderings of analysed minutes.

\subsubsection*{2.7.3 Sensitivity analyses}
We varied (i) the BirdNET confidence threshold from 0.10 to 0.90 in steps of 0.05, (ii) the time-bin width (15, 30, 60 and 120 min) and (iii) the location-filter threshold (0.03, 0.05, 0.10, 0.15, 0.20 and 0.30). For the location filter, S\textsubscript{ref} was recomputed at each threshold.

\subsubsection*{2.7.4 Simulation study}
A simulation tested what each index responds to. Communities of S species were drawn from a pool of 100 species. Abundances followed a log-normal distribution with standard deviation $\sigma$ (unevenness). Each species' activity in 24 time bins followed a gamma distribution with shape k, so that its true temporal CV is 1/$\surd$k. Detections were Poisson with an expected total of N\textsubscript{tot} = 3,000 over n = 24 bins, with the abundances rescaled to sum to N\textsubscript{tot}:

\begin{equation}\lambda_i \sim \mathrm{LogNormal}(0, \sigma^2), \quad \textstyle\sum_i \lambda_i = N_{\mathrm{tot}}; \qquad a_{i,t} \sim \mathrm{Gamma}(k, 1/k); \qquad c_{i,t} \sim \mathrm{Poisson}\!\left(\lambda_i\, a_{i,t} / n\right)\tag{17}\end{equation}
False positives were added as a fraction of true detections, spread over random pool species and bins. Four scenarios varied one factor at a time from a baseline of S = 40, $\sigma$ = 1, k = 4 and no false positives: richness (5–100), unevenness ($\sigma$ = 0–3), temporal CV (0.1–1.5) and false-positive rate (0–30\%). Each scenario used 40 replicates.

\subsubsection*{2.7.5 Hour-level analyses}
Within each recorded hour we computed richness, Shannon H, ABI\textsubscript{E} and ABI\textsubscript{rev} (with 10-min sub-bins for T) and the hourly mean of each acoustic index and of anthropophony. Relationships were measured with Spearman's rank correlation. Each zoo site was contrasted with Chirundu with Mann–Whitney U tests, and p-values were adjusted with the Holm method across all metrics and sites (Holm, 1979). Effect sizes were expressed as Cliff's $\delta$ (Cliff, 1993), the difference between the probabilities that a Chirundu hour a exceeds a zoo hour b and the reverse, over all n\textsubscript{a} n\textsubscript{b} pairs:

\begin{equation}\delta = \frac{\#\{(a,b): a > b\} - \#\{(a,b): a < b\}}{n_a\, n_b}\tag{18}\end{equation}
\subsection*{2.8 Software and reproducibility}
Analyses used Python 3 with PyTorch and torchvision (CNNs), scikit-learn (random forest, logistic regression, cross-validation), librosa (audio features), scikit-maad (acoustic indices), birdnetlib and the BirdNET v2.4 TFLite model (detection and location filter), pandas, SciPy and statsmodels (statistics) and matplotlib and cartopy (figures). The CNNs were trained on an NVIDIA T4 GPU in Google Colab. All random seeds were fixed. The code is organised as a Python package with one command per step (benchmark, field, analysis), which reproduces every number, table and figure in this paper from the raw recordings.

\section*{3. Results}
\subsection*{3.1 Benchmark data and classifier performance}
The RFCx benchmark contained 1,216 calls in 24 classes, with between 34 and 100 calls per class (Table 2; Fig. 3). Under recording-grouped cross-validation, BirdNET embeddings + logistic regression achieved the highest macro-F1 (0.750 $\pm$ 0.029) and LWLRAP (0.846 $\pm$ 0.017) (Table 3; Fig. 4). MobileNetV2 (0.738) and EfficientNet-B0 (0.737) were close behind, whereas the random forest (MFCC statistics) was clearly weaker (0.555). The embedding model needed no GPU training. Calibration was best for MobileNetV2 (ECE = 0.079) and slightly poorer for the embedding model (ECE = 0.108; Fig. S1). For both CNNs, inner-validation macro-F1 rose steeply over the first 10 epochs and levelled off at about 0.72–0.79 after about 20 epochs, with a modest gap between training and validation loss (Fig. 5). Per-class recall of the best model ranged from 0.60 to 0.94 for 23 of the 24 classes; class 3 was the exception (0.32), with its errors spread over several other classes (Fig. 6; Table 4; Fig. 7).

{\small\setlength{\tabcolsep}{4pt}
\begin{longtable}{lccccc}
\caption{\textbf{Table 2.} RFCx benchmark data: annotations, recordings and median call duration and frequency range per class.}\\
\toprule
\textbf{Class} & \textbf{Calls} & \textbf{Recordings} & \textbf{Median duration (s)} & \textbf{Median f\textsubscript{min} (Hz)} & \textbf{Median f\textsubscript{max} (Hz)} \\
\midrule
\endfirsthead
\toprule
\textbf{Class} & \textbf{Calls} & \textbf{Recordings} & \textbf{Median duration (s)} & \textbf{Median f\textsubscript{min} (Hz)} & \textbf{Median f\textsubscript{max} (Hz)} \\
\midrule
\endhead
\bottomrule
\endlastfoot
0 & 50 & 50 & 0.79 & 5,906 & 8,250 \\
1 & 50 & 50 & 0.90 & 3,844 & 5,625 \\
2 & 37 & 30 & 1.78 & 469 & 3,000 \\
3 & 50 & 50 & 0.74 & 1,031 & 2,812 \\
4 & 50 & 50 & 1.86 & 2,344 & 4,219 \\
5 & 50 & 50 & 2.35 & 4,781 & 10,406 \\
6 & 50 & 49 & 2.14 & 562 & 4,406 \\
7 & 50 & 50 & 3.14 & 4,688 & 11,438 \\
8 & 50 & 50 & 4.12 & 3,750 & 5,531 \\
9 & 34 & 33 & 3.56 & 844 & 5,719 \\
10 & 50 & 49 & 2.44 & 947 & 10,853 \\
11 & 50 & 50 & 1.14 & 1,809 & 5,685 \\
12 & 50 & 50 & 3.14 & 562 & 3,281 \\
13 & 50 & 50 & 0.62 & 94 & 844 \\
14 & 50 & 50 & 0.59 & 2,812 & 5,531 \\
15 & 50 & 47 & 3.34 & 94 & 1,125 \\
16 & 50 & 43 & 1.83 & 3,094 & 4,594 \\
17 & 68 & 50 & 6.45 & 1,312 & 3,938 \\
18 & 50 & 50 & 0.69 & 3,188 & 5,062 \\
19 & 43 & 41 & 1.55 & 281 & 2,812 \\
20 & 40 & 35 & 3.44 & 2,344 & 5,719 \\
21 & 50 & 50 & 1.09 & 3,281 & 4,125 \\
22 & 44 & 44 & 1.61 & 10,688 & 13,688 \\
23 & 100 & 90 & 7.00 & 6,848 & 11,456 \\
\end{longtable}
}

\begin{figure}[!htbp]\centering
\includegraphics[width=\textwidth,height=0.78\textheight,keepaspectratio]{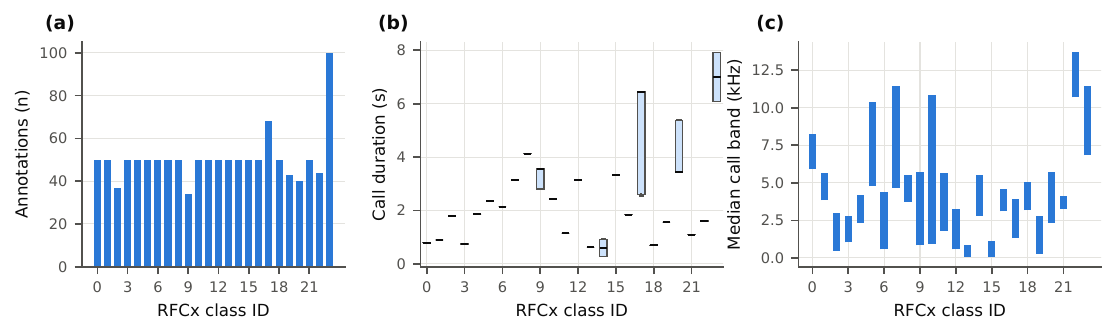}\par\medskip
\caption{\textbf{Fig. 3.} The RFCx strongly labelled benchmark: number of annotated calls per class and the frequency range and duration of the calls.}
\end{figure}

{\footnotesize\setlength{\tabcolsep}{4pt}
\begin{longtable}{lcccccc}
\caption{\textbf{Table 3.} Classifier performance under stratified, recording-grouped 5-fold cross-validation (mean $\pm$ s.d. over folds). ECE = expected calibration error (lower is better).}\\
\toprule
\textbf{Model} & \textbf{Accuracy} & \textbf{Balanced acc.} & \textbf{Macro-F1} & \textbf{Top-3 acc.} & \textbf{LWLRAP} & \textbf{ECE} \\
\midrule
\endfirsthead
\toprule
\textbf{Model} & \textbf{Accuracy} & \textbf{Balanced acc.} & \textbf{Macro-F1} & \textbf{Top-3 acc.} & \textbf{LWLRAP} & \textbf{ECE} \\
\midrule
\endhead
\bottomrule
\endlastfoot
BirdNET emb. + LR & 0.754 $\pm$ 0.024 & 0.757 $\pm$ 0.026 & 0.750 $\pm$ 0.029 & 0.933 $\pm$ 0.010 & 0.846 $\pm$ 0.017 & 0.108 $\pm$ 0.024 \\
MobileNetV2 & 0.749 $\pm$ 0.015 & 0.752 $\pm$ 0.015 & 0.738 $\pm$ 0.020 & 0.917 $\pm$ 0.011 & 0.837 $\pm$ 0.011 & 0.079 $\pm$ 0.020 \\
EfficientNet-B0 & 0.754 $\pm$ 0.013 & 0.751 $\pm$ 0.015 & 0.737 $\pm$ 0.015 & 0.905 $\pm$ 0.009 & 0.838 $\pm$ 0.008 & 0.083 $\pm$ 0.015 \\
Random forest (MFCC) & 0.583 $\pm$ 0.040 & 0.571 $\pm$ 0.028 & 0.555 $\pm$ 0.029 & 0.817 $\pm$ 0.026 & 0.720 $\pm$ 0.027 & 0.309 $\pm$ 0.035 \\
\end{longtable}
}

\begin{figure}[!htbp]\centering
\includegraphics[width=\textwidth,height=0.78\textheight,keepaspectratio]{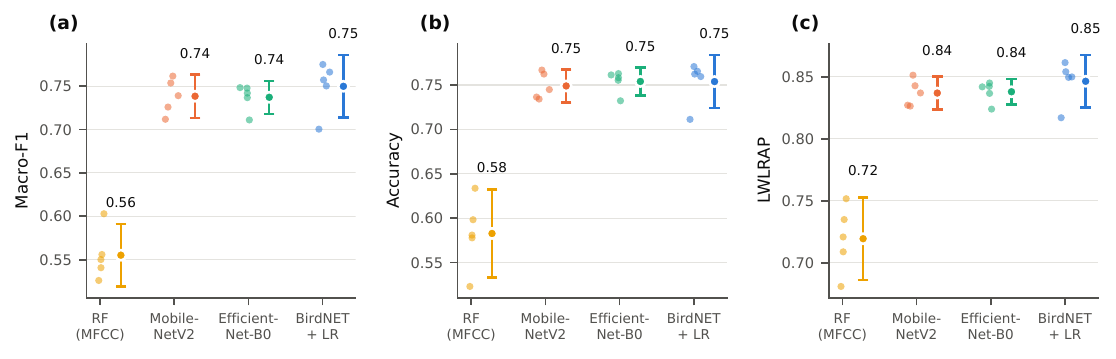}\par\medskip
\caption{\textbf{Fig. 4.} Comparison of the four classifiers under recording-grouped 5-fold cross-validation (mean and 95\% confidence interval over folds).}
\end{figure}

\begin{figure}[!htbp]\centering
\includegraphics[width=\textwidth,height=0.78\textheight,keepaspectratio]{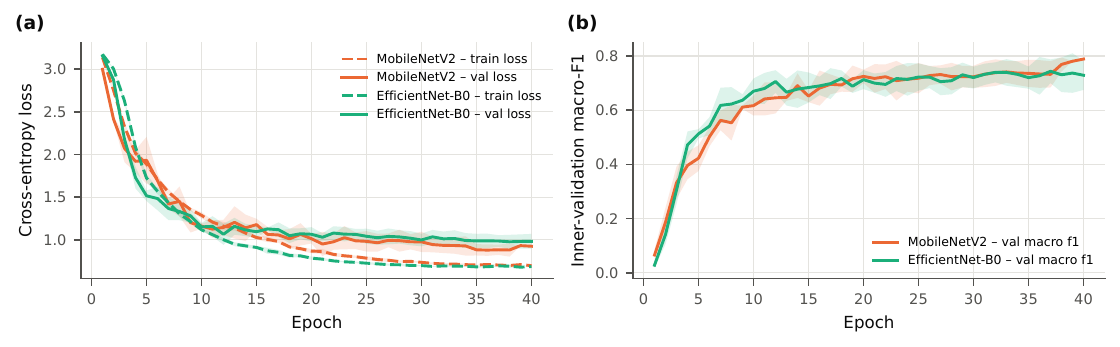}\par\medskip
\caption{\textbf{Fig. 5.} Training loss and inner-validation macro-F1 of MobileNetV2 and EfficientNet-B0 by epoch for each fold; training stopped when validation macro-F1 had not improved for 10 epochs.}
\end{figure}

\begin{figure}[!htbp]\centering
\includegraphics[width=\textwidth,height=0.78\textheight,keepaspectratio]{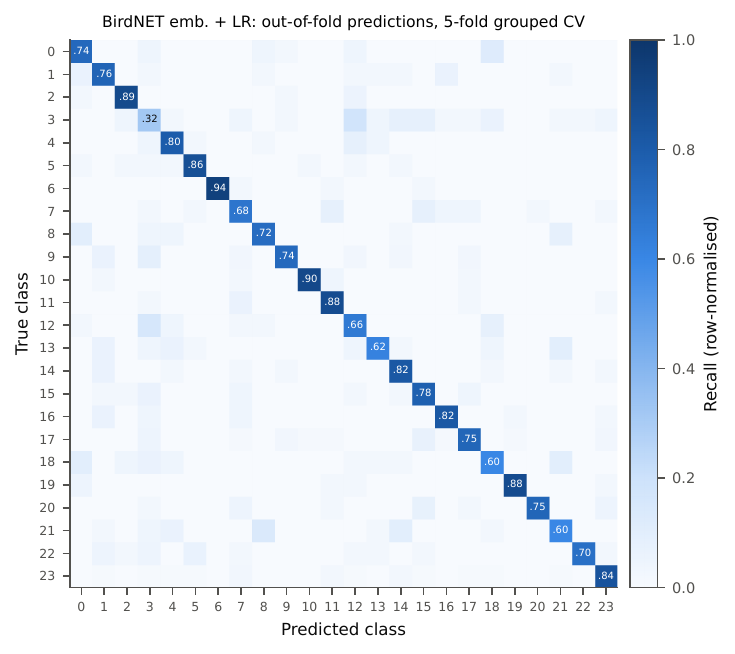}\par\medskip
\caption{\textbf{Fig. 6.} Row-normalised confusion matrix of the best model (BirdNET embeddings + logistic regression), pooled over the out-of-fold predictions of the five folds.}
\end{figure}

{\footnotesize\setlength{\tabcolsep}{4pt}
\begin{longtable}{lcccc}
\caption{\textbf{Table 4.} Per-class F1 score of each model (pooled out-of-fold predictions).}\\
\toprule
\textbf{Class} & \textbf{BirdNET emb. + LR} & \textbf{EfficientNet-B0} & \textbf{MobileNetV2} & \textbf{Random forest (MFCC)} \\
\midrule
\endfirsthead
\toprule
\textbf{Class} & \textbf{BirdNET emb. + LR} & \textbf{EfficientNet-B0} & \textbf{MobileNetV2} & \textbf{Random forest (MFCC)} \\
\midrule
\endhead
\bottomrule
\endlastfoot
0 & 0.70 & 0.74 & 0.78 & 0.59 \\
1 & 0.72 & 0.70 & 0.71 & 0.66 \\
2 & 0.86 & 0.83 & 0.87 & 0.74 \\
3 & 0.30 & 0.08 & 0.17 & 0.22 \\
4 & 0.75 & 0.70 & 0.71 & 0.73 \\
5 & 0.86 & 0.86 & 0.87 & 0.65 \\
6 & 0.97 & 0.92 & 0.91 & 0.75 \\
7 & 0.65 & 0.55 & 0.54 & 0.42 \\
8 & 0.73 & 0.72 & 0.73 & 0.74 \\
9 & 0.77 & 0.71 & 0.74 & 0.11 \\
10 & 0.93 & 0.95 & 0.91 & 0.70 \\
11 & 0.83 & 0.78 & 0.73 & 0.76 \\
12 & 0.60 & 0.63 & 0.61 & 0.53 \\
13 & 0.70 & 0.91 & 0.90 & 0.51 \\
14 & 0.75 & 0.68 & 0.71 & 0.69 \\
15 & 0.72 & 0.74 & 0.75 & 0.57 \\
16 & 0.84 & 0.84 & 0.80 & 0.76 \\
17 & 0.79 & 0.88 & 0.90 & 0.49 \\
18 & 0.61 & 0.50 & 0.48 & 0.53 \\
19 & 0.92 & 0.89 & 0.89 & 0.62 \\
20 & 0.85 & 0.92 & 0.89 & 0.44 \\
21 & 0.61 & 0.56 & 0.53 & 0.45 \\
22 & 0.82 & 0.83 & 0.84 & 0.39 \\
23 & 0.86 & 0.90 & 0.88 & 0.53 \\
\end{longtable}
}

\begin{figure}[!htbp]\centering
\includegraphics[width=\textwidth,height=0.78\textheight,keepaspectratio]{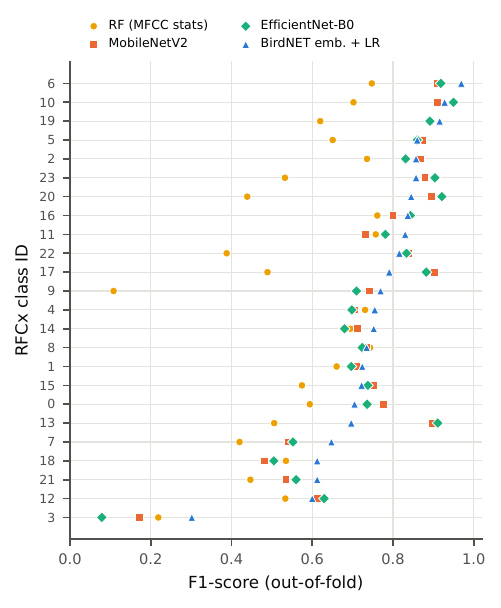}\par\medskip
\caption{\textbf{Fig. 7.} Per-class F1 score of the four classifiers.}
\end{figure}

Evaluation protocol had a large effect. In the frame-level analysis, a random split of frames gave an accuracy of 0.70, whereas grouping by recording gave 0.41 (Table S1): about 29 percentage points of the apparent performance came from frames of the same calls appearing in training and test sets. The RFCx-trained CNN did not transfer to Zimbabwe. Its median maximum softmax probability was 0.82 on RFCx out-of-fold predictions but 0.19 on 720 Chirundu windows, and only 0.1\% of Chirundu windows exceeded 0.5 (Table S2; Fig. S2). The field analysis therefore used BirdNET's named species with a location filter rather than any RFCx-trained model.

\subsection*{3.2 Soundscape structure}
The false-colour spectrograms show a very different structure at each site (Fig. 8). At Chirundu, energy between 2 and 8 kHz was structured into many discrete events, with narrow-band insect choruses near 7–9 kHz and broadband bird activity from dawn to late morning. At Cage and Walking, a broadband source was present from about 06:00 to 18:00 every day and absent at night, when narrow-band insect choruses near 7–8 kHz dominated. The source of this daytime noise was not identified; ventilation, misting and water features are likely candidates. The Pool had near-constant broadband noise day and night, consistent with water pumps.

These sources shaped the diel profiles of the indices (Fig. 9). At Cage and Walking, ADI and acoustic entropy H switched to near their maximum values at about 06:00 and back at about 18:00, and AEI did the opposite. At the Pool, ADI and H stayed high throughout. Chirundu had the highest ACI in most recorded hours, and its ADI, AEI and H lay between the zoo's daytime and night-time values. Example spectrograms with BirdNET detections are shown in Fig. 10.

\begin{figure}[!htbp]\centering
\includegraphics[width=\textwidth,height=0.78\textheight,keepaspectratio]{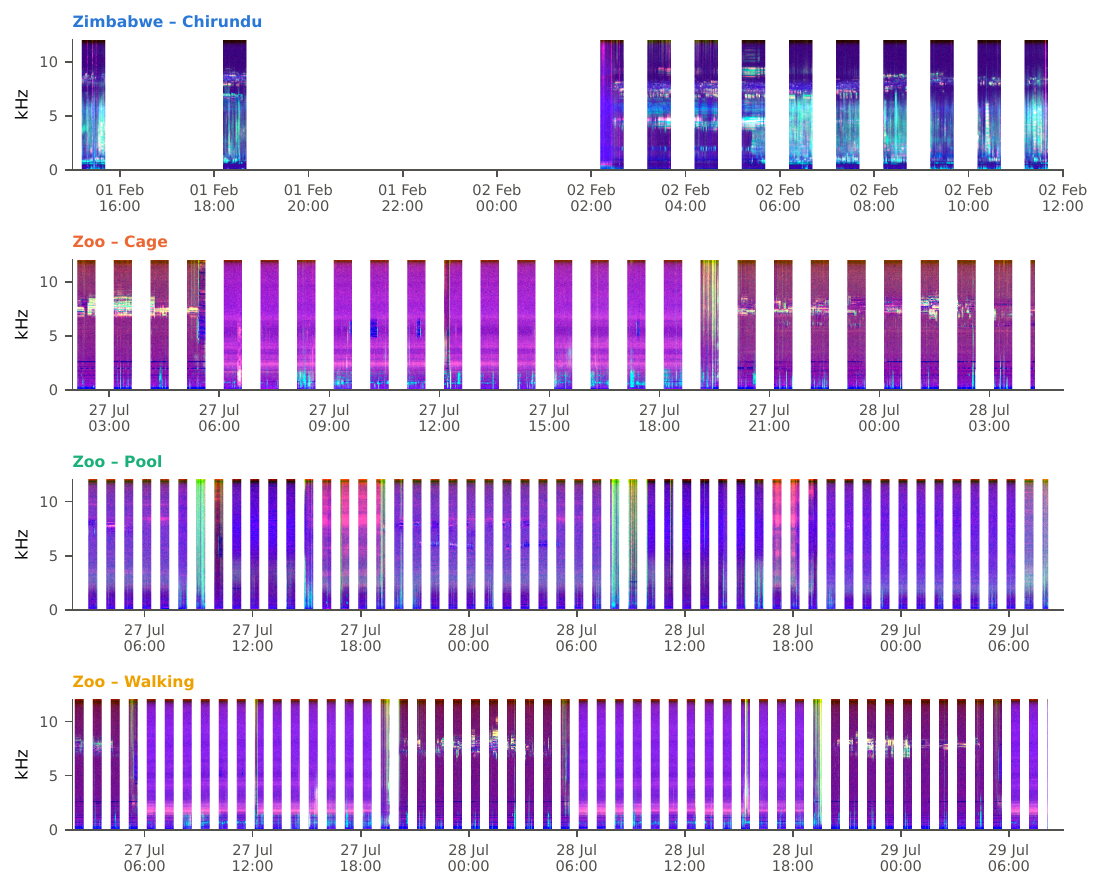}\par\medskip
\caption{\textbf{Fig. 8.} Long-duration false-colour spectrograms of the four sites (0–12 kHz). Red = acoustic complexity, green = event fraction (energy > background + 6 dB), blue = mean spectral power, each percentile-normalised per site. White = not recorded.}
\end{figure}

\begin{figure}[!htbp]\centering
\includegraphics[width=\textwidth,height=0.78\textheight,keepaspectratio]{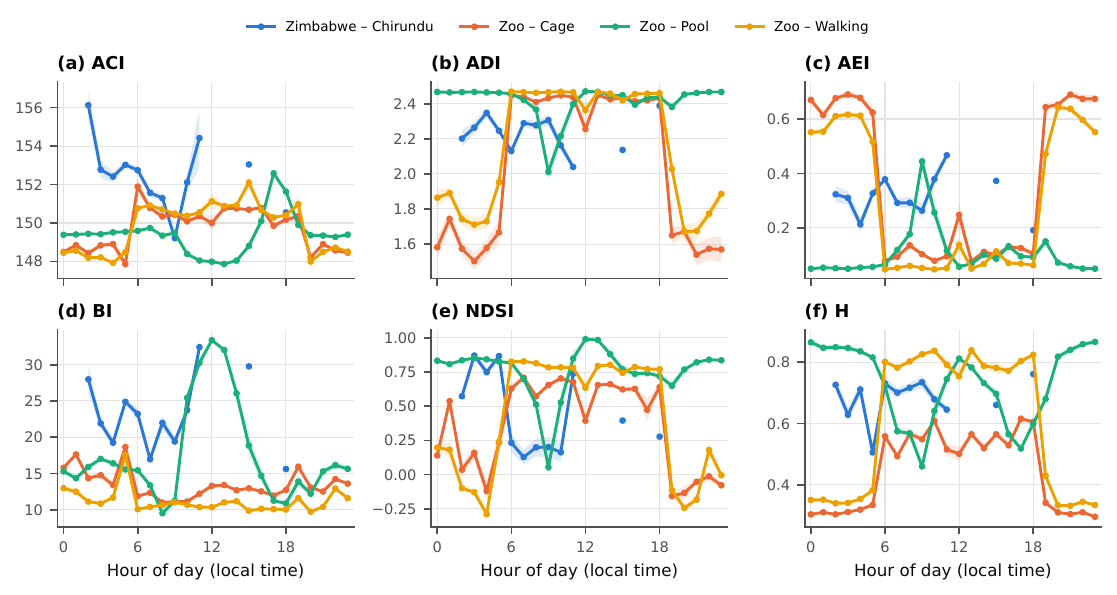}\par\medskip
\caption{\textbf{Fig. 9.} Diel profiles of the six acoustic indices and the sound level at each site (hourly means of 1-min values).}
\end{figure}

\begin{figure}[!htbp]\centering
\includegraphics[width=\textwidth,height=0.78\textheight,keepaspectratio]{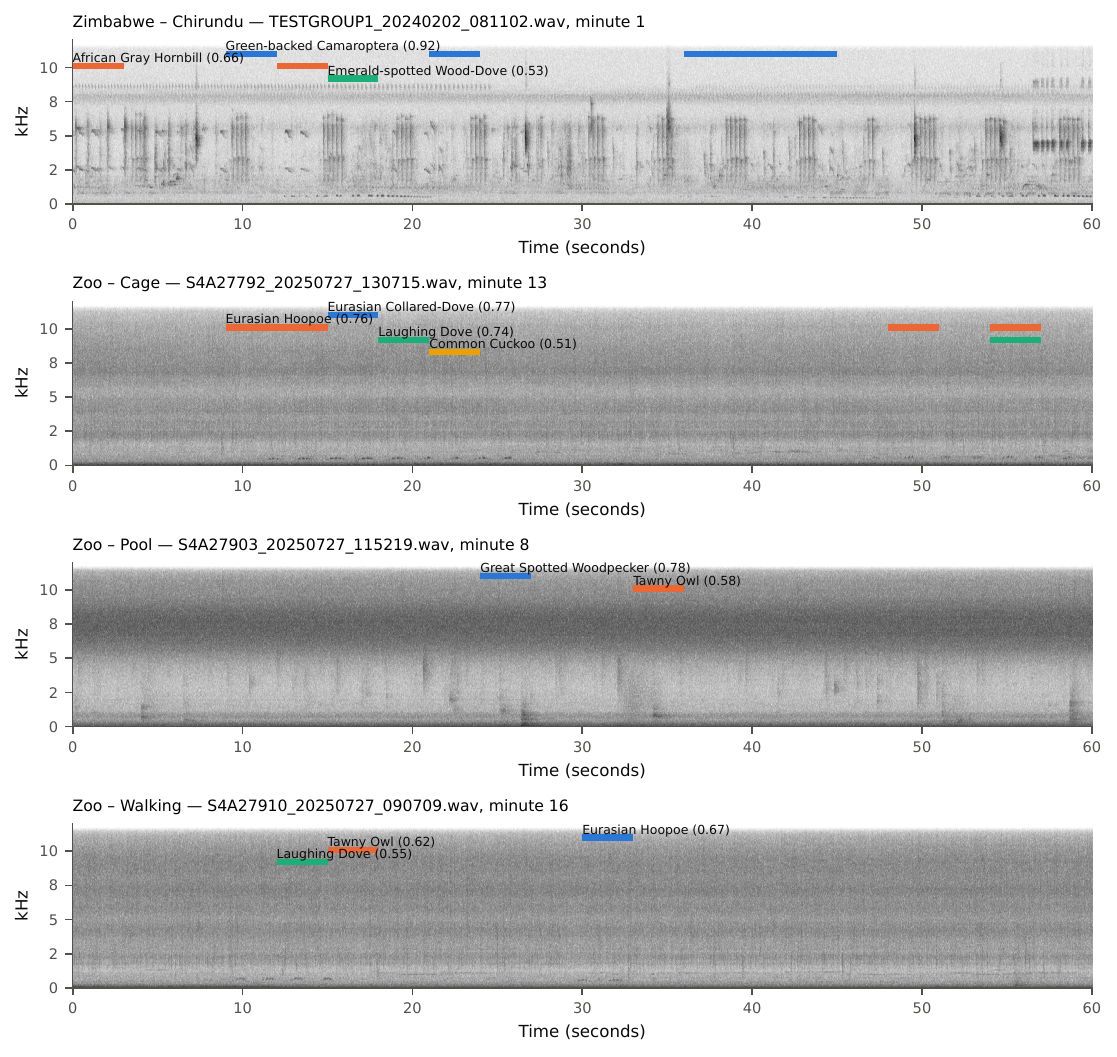}\par\medskip
\caption{\textbf{Fig. 10.} Example spectrograms from each site with BirdNET detections marked.}
\end{figure}

\subsection*{3.3 Detections and false-positive control}
The strict location filter removed 517 detections of 9 species (13 site–species records), all at the zoo; no Chirundu detection was removed (Table S8). Tawny Owl (\textit{Strix aluco}) accounted for 466 of the removed detections (Pool 135, Walking 307, Cage 24), with confidence up to 0.999. Great Spotted Woodpecker (\textit{Dendrocopos major}) added 42 more. Both are Palearctic woodland species without plausible wild occurrence in Abu Dhabi in July. The remaining removals were single detections of other Palearctic species. These detections are therefore most likely captive animals or non-avian sounds that BirdNET classified with high confidence.

After filtering, Chirundu held 287 detections of 21 species. The most frequent were Green-backed Camaroptera (\textit{Camaroptera brachyura}, 94), Yellow-bellied Greenbul (\textit{Chlorocichla flaviventris}, 29), Black-backed Puffback (\textit{Dryoscopus cubla}, 25), Ring-necked Dove (\textit{Streptopelia capicola}, 24), Arrow-marked Babbler (\textit{Turdoides jardineii}, 23) (Table S11; Fig. 11). The zoo sites held 459 detections of 17 species (Cage), 678 of 25 (Walking) and only 9 of 4 (Pool). The most frequent zoo species were Cage: Laughing Dove (317), Eurasian Collared-Dove (60), Eurasian Hoopoe (24); Walking: Laughing Dove (431), Eurasian Collared-Dove (73), House Crow (53); Pool: Mallard (6), Dunlin (1), Eurasian Collared-Dove (1). A few species that passed the filter are still unexpected in Abu Dhabi in July, for example Eurasian Wigeon, Great Bittern and Tree Pipit, each with few detections; they may be captive birds or misclassifications and are priorities for expert verification. These species were identified by screening the filtered lists against the known status of birds in the United Arab Emirates. They were retained in the metrics because removing them changes ABI\textsubscript{E} by less than 0.005 at Cage and Walking, and their exported clips are included in the verification set. Detection rates followed a strong diel pattern at Chirundu: there were almost no detections in the recorded pre-dawn hours (02:00–05:00), and most were made between 06:00 and 11:00, peaking at 08:00 and 10:00–11:00. Zoo detections were spread more evenly across the day and night (Fig. 12).

\begin{figure}[!htbp]\centering
\includegraphics[width=\textwidth,height=0.78\textheight,keepaspectratio]{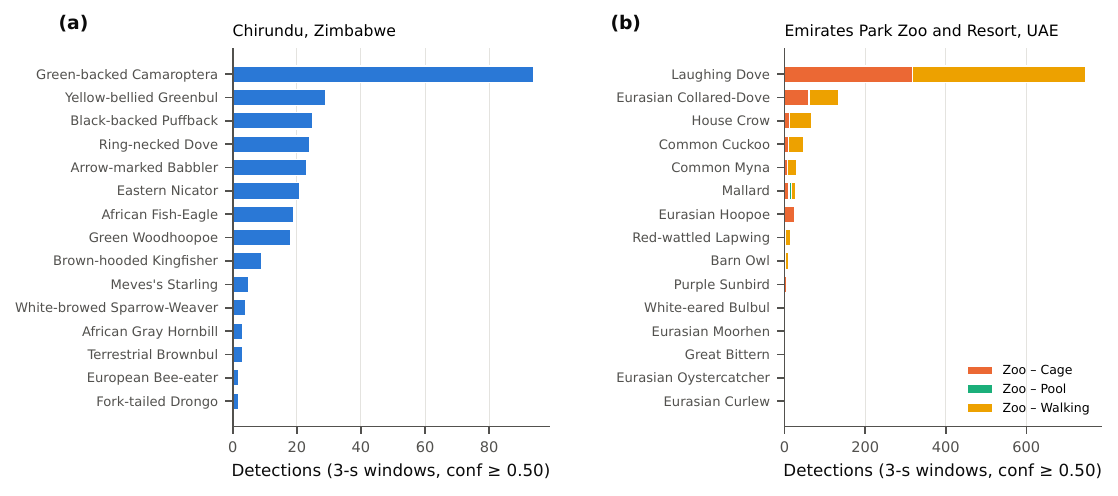}\par\medskip
\caption{\textbf{Fig. 11.} Species composition: the 15 most frequently detected species at (a) Chirundu and (b) the zoo, stacked by recorder (3-s windows with confidence $\geq$ 0.50, after the strict location filter).}
\end{figure}

\begin{figure}[!htbp]\centering
\includegraphics[width=\textwidth,height=0.78\textheight,keepaspectratio]{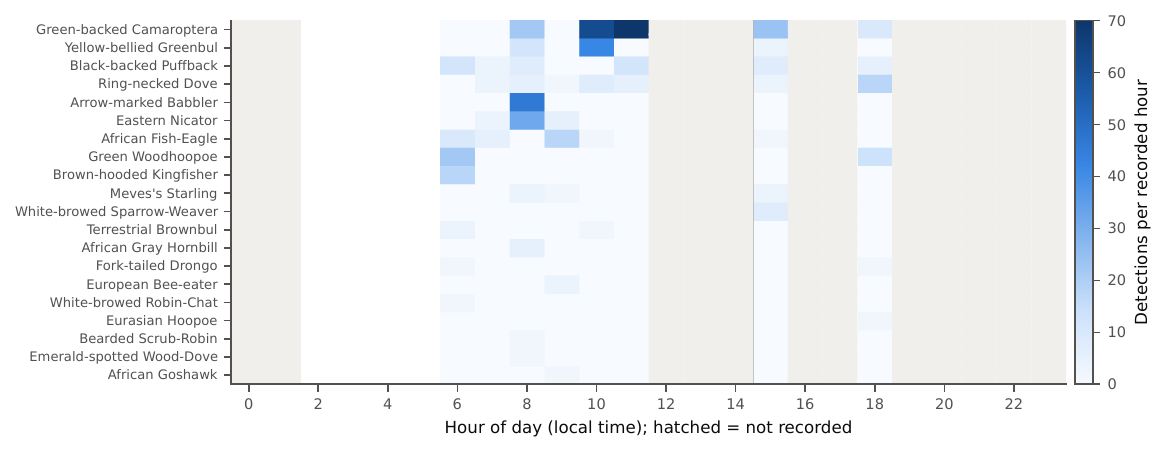}\par\medskip
\includegraphics[width=\textwidth,height=0.78\textheight,keepaspectratio]{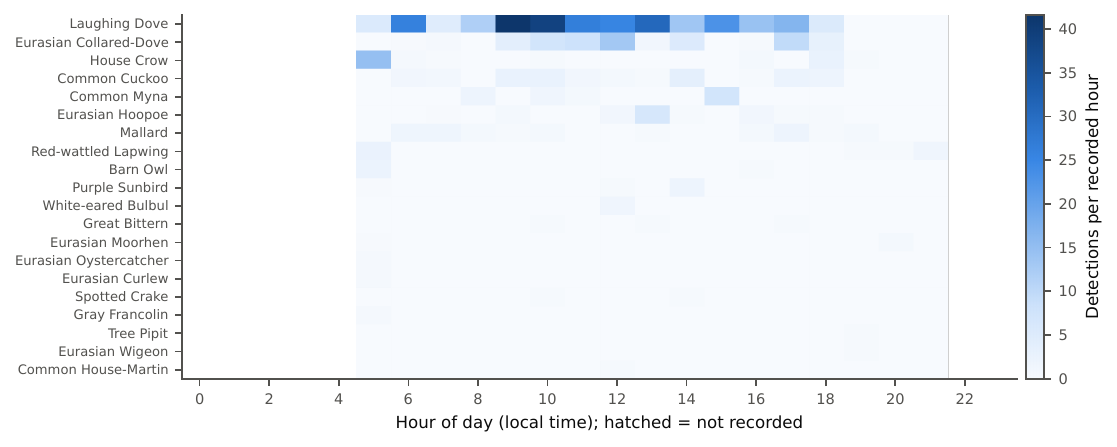}\par\medskip
\caption{\textbf{Fig. 12.} Diel activity of the 20 most frequent species at (top) Chirundu and (bottom) the zoo, as detections per recorded hour. Hatched columns were not recorded.}
\end{figure}

\subsection*{3.4 Site comparison at full effort}
Chirundu had the highest effective diversity (\textsuperscript{1}D = 9.83 vs 2.73–4.26 at the zoo) and the highest ABI\textsubscript{E} (Eq. 11; 0.38 [0.25–0.41]), although it had the least effort (Table 5; Fig. 13). Walking had more species (25 vs 21) with 4.5 times the effort, but a much lower \textsuperscript{1}D because a few species dominated its detections (J$'$ = 0.45). The Pool, with 9 bird detections in 26.8 h, was almost devoid of birds.

The combined index ABI\textsubscript{rev} (Eq. 14) did not separate Chirundu (0.150 [0.096–0.214]) from Walking (0.162 [0.129–0.189]) or Cage (0.132 [0.096–0.159]). The reason is the stability term: T was higher at Walking (0.67) and Cage (0.65) than at Chirundu (0.39), because Chirundu's birds were silent in the pre-dawn hours and concentrated their calling in the morning. The original ABI (Eq. 15) ranked Chirundu highest (0.205), but only because its evenness J$'$ was high; it is insensitive to richness (Section 3.7).

{\footnotesize\setlength{\tabcolsep}{4pt}
\begin{longtable}{lcccccccccc}
\caption{\textbf{Table 5.} Site-level diversity and ABI at full effort (60-min bins, confidence $\geq$ 0.50, strict location filter). Brackets: 95\% block-bootstrap confidence intervals.}\\
\toprule
\textbf{Site} & \textbf{Effort (h)} & \textbf{Detections} & \textbf{S} & \textbf{\textsuperscript{1}D} & \textbf{\textsuperscript{2}D} & \textbf{J$'$} & \textbf{ABI\textsubscript{E}} & \textbf{T} & \textbf{ABI\textsubscript{rev}} & \textbf{ABI\textsubscript{orig}} \\
\midrule
\endfirsthead
\toprule
\textbf{Site} & \textbf{Effort (h)} & \textbf{Detections} & \textbf{S} & \textbf{\textsuperscript{1}D} & \textbf{\textsuperscript{2}D} & \textbf{J$'$} & \textbf{ABI\textsubscript{E}} & \textbf{T} & \textbf{ABI\textsubscript{rev}} & \textbf{ABI\textsubscript{orig}} \\
\midrule
\endhead
\bottomrule
\endlastfoot
Chirundu & 6.0 & 287 & 21 & 9.83 & 6.49 & 0.75 & 0.38 [0.25–0.41] & 0.39 & 0.150 [0.096–0.214] & 0.205 \\
Zoo – Walking & 27.0 & 678 & 25 & 4.26 & 2.34 & 0.45 & 0.24 [0.17–0.29] & 0.67 & 0.162 [0.129–0.189] & 0.069 \\
Zoo – Cage & 13.1 & 459 & 17 & 3.37 & 2.00 & 0.43 & 0.20 [0.14–0.25] & 0.65 & 0.132 [0.096–0.159] & 0.088 \\
Zoo – Pool & 26.8 & 9 & 4 & 2.73 & 2.08 & 0.72 & 0.17 [0.00–0.21] & 0.36 & 0.060 [0.000–0.090] & 0.098 \\
\end{longtable}
}

\begin{figure}[!htbp]\centering
\includegraphics[width=\textwidth,height=0.78\textheight,keepaspectratio]{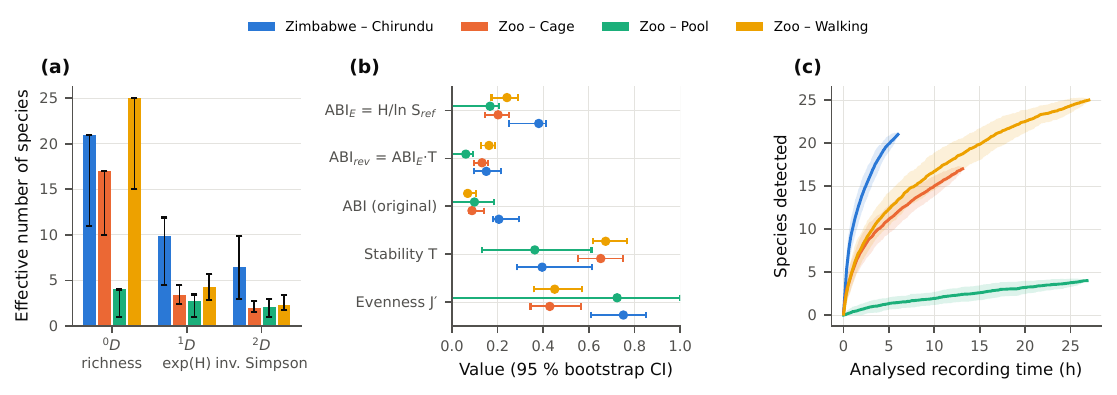}\par\medskip
\caption{\textbf{Fig. 13.} Site comparison at full effort. (a) Hill numbers \textsuperscript{0}D (richness), \textsuperscript{1}D and \textsuperscript{2}D with 95\% block-bootstrap confidence intervals. (b) ABI and its components. (c) Species accumulation with analysed recording time (mean $\pm$ s.d. of 100 random orderings of minutes).}
\end{figure}

\subsection*{3.5 Effort-standardised comparison and robustness}
Diel-matched rarefaction to Chirundu's effort made the difference between sites larger (Table S9; Fig. S3a–b). Chirundu kept S = 21 and ABI\textsubscript{E} = 0.38, whereas the zoo sites fell to S = 13.0 (Cage), 12.6 [10.0–15.0] (Walking) and 1.3 [1.0–2.0] (Pool), with \textsuperscript{1}D = 2.93, 3.51 and 1.28 and ABI\textsubscript{E} = 0.18, 0.21 [0.16–0.25] and 0.03 [0.00–0.12]. The Cage values have no spread because that recorder had exactly one bin for each hour of the day sampled at Chirundu, so every draw selected the same bins.

Chirundu had the highest ABI\textsubscript{E} at every threshold of the location filter (0.03–0.30; Table S10; Fig. S3c). Chirundu was almost unaffected (S = 21 throughout; ABI\textsubscript{E} = 0.37–0.39). Zoo values fell as the filter tightened (Cage S from 19 to 11, ABI\textsubscript{E} from 0.22 to 0.18; Walking S from 32 to 20, ABI\textsubscript{E} from 0.27 to 0.21), as expected if the removed species were false positives.

Chirundu had the highest ABI\textsubscript{E} at 17 of 17 confidence thresholds between 0.10 and 0.90 (Table S4; Fig. 14a–b). Its richness fell from 62 species at the lowest threshold to 10 at the highest. ABI\textsubscript{E} does not depend on bin width, because H and S\textsubscript{ref} do not. Changing the bin width from 15 to 120 min changed T by up to 0.09 and ABI\textsubscript{rev} by up to 0.04 (Table S5; Fig. 14c).

\begin{figure}[!htbp]\centering
\includegraphics[width=\textwidth,height=0.78\textheight,keepaspectratio]{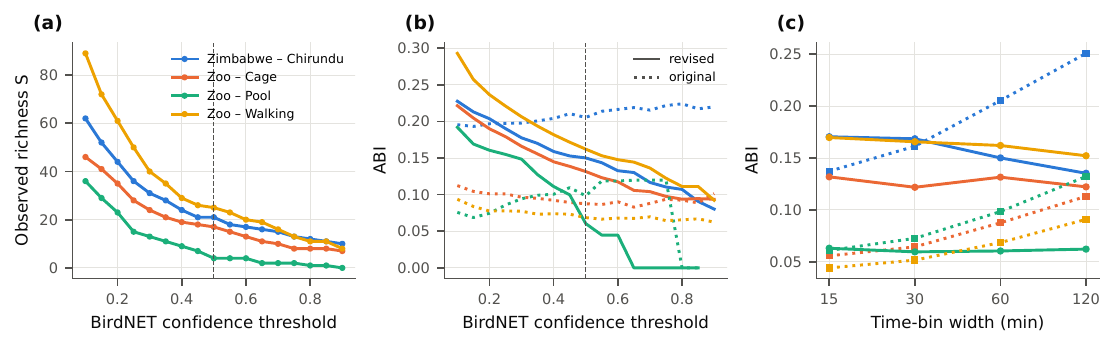}\par\medskip
\caption{\textbf{Fig. 14.} Sensitivity of the indices to analysis choices. (a) Observed richness and (b) ABI\textsubscript{rev} (solid) and ABI\textsubscript{orig} (dotted) against the BirdNET confidence threshold; the dashed line marks the threshold used. (c) ABI against time-bin width.}
\end{figure}

\subsection*{3.6 Relationship with standard acoustic indices}
Across 149 recorded hours, ABI\textsubscript{E} tracked BirdNET richness closely ($\rho$ = 0.93), whereas no standard index exceeded |$\rho$| = 0.48 (Table 6; Fig. 15). ACI was the best index ($\rho$ = 0.48). ADI, AEI, NDSI and H were essentially uncorrelated with richness (|$\rho$| $\leq$ 0.16), and BI was negatively correlated ($\rho$ = $-$0.26). Anthropophony was weakly positively correlated with richness ($\rho$ = 0.26), because visitor activity and bird activity both peak in daylight. Within this analysis ABI\textsubscript{E} and Shannon H had identical ranks ($\rho$ = 1.00), because S\textsubscript{ref} is a constant.

Hour-by-hour contrasts with Chirundu (12 hours) were significant for richness and ABI\textsubscript{E} only against the Pool (Cliff's $\delta$ = 0.62 and 0.66; Holm-adjusted p < 0.001 and p < 0.001; Table 7). Against Cage and Walking the effects on ABI\textsubscript{E} were moderate ($\delta$ = 0.40 and 0.39) but not significant (Holm-adjusted p = 0.568 and p = 0.476). ACI and BI were higher at Chirundu than at every zoo site (ACI: $\delta$ = 0.86, 0.82, 0.84; BI: $\delta$ = 0.94, 0.98, 0.56), and anthropophony was higher at all zoo sites ($\delta$ = $-$0.70, $-$0.60, $-$0.53). ADI and NDSI were highest at the Pool, the site with almost no bird detections.

{\small\setlength{\tabcolsep}{4pt}
\begin{longtable}{lcccc}
\caption{\textbf{Table 6.} Spearman rank correlations between hourly metrics (149 recorded hours, four sites).}\\
\toprule
\textbf{Metric} & \textbf{$\rho$ with richness} & \textbf{$\rho$ with Shannon H} & \textbf{$\rho$ with ABI\textsubscript{E}} & \textbf{$\rho$ with anthropophony} \\
\midrule
\endfirsthead
\toprule
\textbf{Metric} & \textbf{$\rho$ with richness} & \textbf{$\rho$ with Shannon H} & \textbf{$\rho$ with ABI\textsubscript{E}} & \textbf{$\rho$ with anthropophony} \\
\midrule
\endhead
\bottomrule
\endlastfoot
ACI & 0.48 & 0.47 & 0.47 & 0.11 \\
ADI & 0.02 & 0.04 & 0.04 & $-$0.06 \\
AEI & $-$0.04 & $-$0.07 & $-$0.07 & 0.03 \\
BI & $-$0.26 & $-$0.21 & $-$0.21 & $-$0.32 \\
NDSI & $-$0.16 & $-$0.11 & $-$0.11 & $-$0.13 \\
H (acoustic entropy) & 0.06 & 0.10 & 0.10 & $-$0.04 \\
Anthropophony & 0.26 & 0.20 & 0.20 & – \\
ABI\textsubscript{rev} & 0.68 & 0.67 & 0.67 & 0.08 \\
ABI\textsubscript{E} & 0.93 & 1.00 & – & 0.20 \\
\end{longtable}
}

{\footnotesize\setlength{\tabcolsep}{4pt}
\begin{longtable}{llcccc}
\caption{\textbf{Table 7.} Hour-level contrasts between Chirundu and each zoo site: medians, Cliff's $\delta$ (positive = higher at Chirundu) and Holm-adjusted p-values of Mann–Whitney U tests.}\\
\toprule
\textbf{Site} & \textbf{Metric} & \textbf{Median Chirundu} & \textbf{Median zoo} & \textbf{Cliff's $\delta$} & \textbf{p (Holm)} \\
\midrule
\endfirsthead
\toprule
\textbf{Site} & \textbf{Metric} & \textbf{Median Chirundu} & \textbf{Median zoo} & \textbf{Cliff's $\delta$} & \textbf{p (Holm)} \\
\midrule
\endhead
\bottomrule
\endlastfoot
Zoo – Cage & Richness & 4.5 & 2 & 0.34 & 0.850 \\
Zoo – Cage & Shannon H & 1.2 & 0.179 & 0.40 & 0.568 \\
Zoo – Cage & ABI\textsubscript{E} & 0.199 & 0.0297 & 0.40 & 0.568 \\
Zoo – Cage & ABI\textsubscript{rev} & 0.0199 & 0 & 0.20 & 1.000 \\
Zoo – Cage & ACI & 153 & 150 & 0.86 & < 0.001 \\
Zoo – Cage & ADI & 2.25 & 1.74 & 0.07 & 1.000 \\
Zoo – Cage & AEI & 0.317 & 0.614 & $-$0.05 & 1.000 \\
Zoo – Cage & BI & 22.6 & 13 & 0.94 & < 0.001 \\
Zoo – Cage & NDSI & 0.337 & 0.475 & 0.23 & 1.000 \\
Zoo – Cage & H (acoustic entropy) & 0.706 & 0.342 & 0.93 & < 0.001 \\
Zoo – Cage & Anthropophony & 0 & 0.0667 & $-$0.70 & 0.004 \\
Zoo – Pool & Richness & 4.5 & 0 & 0.62 & < 0.001 \\
Zoo – Pool & Shannon H & 1.2 & 0 & 0.66 & < 0.001 \\
Zoo – Pool & ABI\textsubscript{E} & 0.199 & 0 & 0.66 & < 0.001 \\
Zoo – Pool & ABI\textsubscript{rev} & 0.0199 & 0 & 0.65 & < 0.001 \\
Zoo – Pool & ACI & 153 & 149 & 0.84 & < 0.001 \\
Zoo – Pool & ADI & 2.25 & 2.46 & $-$0.87 & < 0.001 \\
Zoo – Pool & AEI & 0.317 & 0.0646 & 0.89 & < 0.001 \\
Zoo – Pool & BI & 22.6 & 16 & 0.56 & 0.046 \\
Zoo – Pool & NDSI & 0.337 & 0.814 & $-$0.57 & 0.039 \\
Zoo – Pool & H (acoustic entropy) & 0.706 & 0.793 & $-$0.36 & 0.568 \\
Zoo – Pool & Anthropophony & 0 & 0.0333 & $-$0.53 & 0.032 \\
Zoo – Walking & Richness & 4.5 & 2 & 0.37 & 0.549 \\
Zoo – Walking & Shannon H & 1.2 & 0.41 & 0.39 & 0.476 \\
Zoo – Walking & ABI\textsubscript{E} & 0.199 & 0.0682 & 0.39 & 0.476 \\
Zoo – Walking & ABI\textsubscript{rev} & 0.0199 & 0 & 0.31 & 0.850 \\
Zoo – Walking & ACI & 153 & 150 & 0.82 & < 0.001 \\
Zoo – Walking & ADI & 2.25 & 2.38 & $-$0.04 & 1.000 \\
Zoo – Walking & AEI & 0.317 & 0.152 & 0.05 & 1.000 \\
Zoo – Walking & BI & 22.6 & 10.6 & 0.98 & < 0.001 \\
Zoo – Walking & NDSI & 0.337 & 0.735 & 0.04 & 1.000 \\
Zoo – Walking & H (acoustic entropy) & 0.706 & 0.705 & $-$0.01 & 1.000 \\
Zoo – Walking & Anthropophony & 0 & 0.0333 & $-$0.60 & 0.009 \\
\end{longtable}
}

\begin{figure}[!htbp]\centering
\includegraphics[width=\textwidth,height=0.78\textheight,keepaspectratio]{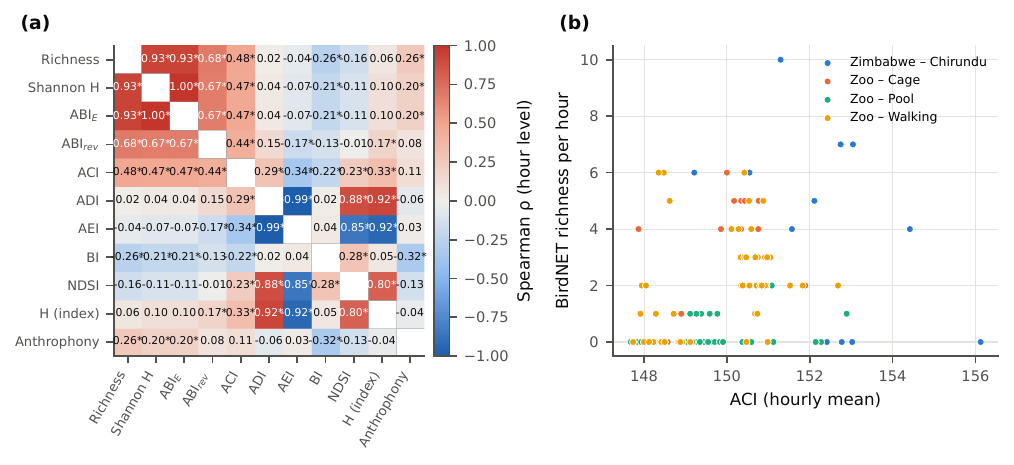}\par\medskip
\caption{\textbf{Fig. 15.} (a) Spearman correlations between hourly metrics (* p < 0.05). (b) Hourly BirdNET richness against the acoustic index most correlated with it (ACI), by site.}
\end{figure}

\subsection*{3.7 Simulation}
The simulation confirmed the analytical properties of each index (Fig. 16; Table S6). As true richness increased from 5 to 100 species, ABI\textsubscript{orig} \textit{decreased} from 0.52 to 0.37 and J$'$ changed little (0.81 to 0.89), whereas ABI\textsubscript{E} increased from 0.28 to 0.89. ABI\textsubscript{E} decreased as abundances became more uneven (0.80 at $\sigma$ = 0 to 0.36 at $\sigma$ = 3). T fell from 0.90 to 0.70 as the true temporal CV rose to 1.5. False positives spread over the pool inflated observed richness from 40 to 100 species and ABI\textsubscript{E} from 0.69 to 0.83 at a false-positive rate of 30\%. This shows why false-positive control must precede any detection-based index.

\begin{figure}[!htbp]\centering
\includegraphics[width=\textwidth,height=0.78\textheight,keepaspectratio]{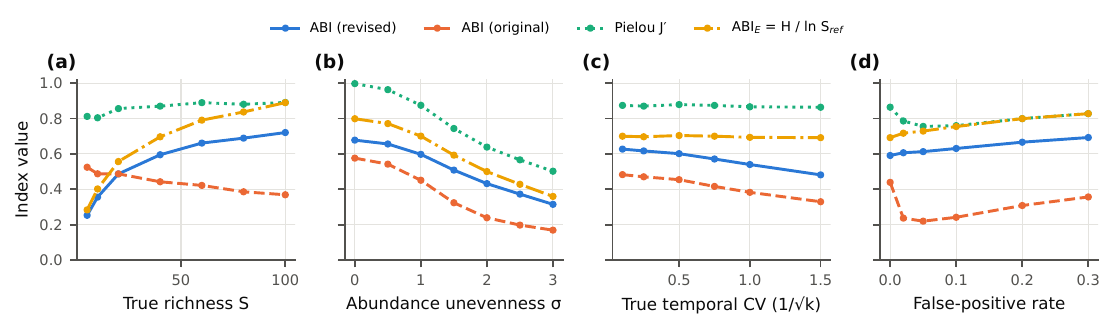}\par\medskip
\caption{\textbf{Fig. 16.} Response of the indices to (a) true richness, (b) abundance unevenness $\sigma$, (c) true temporal CV and (d) the false-positive rate in simulated communities (means of 40 replicates; pool of 100 species, 24 time bins).}
\end{figure}

\section*{4. Discussion}
\subsection*{4.1 Species-level detections outperform energy-based indices}
Species-based metrics separated a natural savanna from three zoo habitats, whereas the six standard acoustic indices did not. At the hourly scale ABI\textsubscript{E} followed richness almost one to one ($\rho$ = 0.93). The best index, ACI, reached 0.48, and BI, ADI, AEI, NDSI and H were near zero or negative. This is consistent with the moderate and context-dependent index–biodiversity relationships reported across studies (Alcocer et al., 2022), and with the biases of indices in urban soundscapes (Fairbrass et al., 2017).

The indices responded to sound sources other than birds. ADI and NDSI were highest at the hippo pool, which held almost no bird detections, and at Cage and Walking they switched with the daytime broadband source rather than with bird activity. Both indices measure how evenly energy fills the frequency bands, and continuous mechanical or water noise fills them well. BI was negatively correlated with richness because it integrates all energy between 2 and 8 kHz, including insects and machinery. The indices still carried information: ACI and BI were higher at Chirundu than at every zoo site. They are best read as descriptors of soundscape structure, not as measures of biodiversity.

The very low diversity at the hippopotamus pool (9 detections of 4 species in 26.8 h) has two likely causes that the index cannot separate. The exhibit is an open water body with little tree or shrub cover, so it offers few perches or foraging sites for the doves, crows and mynas that dominate the other two zoo sites, and the species detected there (Mallard, Dunlin, Barn Owl) are consistent with a water body or captive birds rather than a passerine community. In addition, the near-constant pump noise raises the level a call must exceed to be detected, so quieter species may be present but missed. The Pool value is therefore best read as a lower bound, and it shows that ABI\textsubscript{E} responds to habitat structure within a single site at the scale of tens of metres.

\subsection*{4.2 What the ABI adds, and what it does not}
ABI\textsubscript{E} is Shannon entropy scaled by the logarithm of a fixed regional pool. Within a study where S\textsubscript{ref} is constant it ranks samples exactly as H does. Its value therefore lies in the framework around it rather than in a new formula:

\begin{itemize}
\item it is bounded in [0, 1] on a scale fixed by the regional pool, so values can be compared across sites and seasons, and read as the fraction of the maximum diversity the region could support;
\item unlike Pielou's J$'$, it increases with richness (Fig. 16a);
\item it is computed from species-level detections that have passed explicit, reproducible false-positive control;
\item it is reported with Hill numbers and effort-standardised estimates, so that richness and dominance can be examined separately.
\end{itemize}
The choice of S\textsubscript{ref} matters. We used the largest BirdNET site list at a meta-model score of 0.10. Studies that compare ABI\textsubscript{E} values should declare and share the pool definition, and report ABI\textsubscript{E} together with \textsuperscript{0}D and \textsuperscript{1}D. Because S\textsubscript{ref} depends on the BirdNET meta-model, it changes with the model version, so the version should be reported too.

\subsection*{4.3 Temporal stability should be reported separately}
The stability term behaved opposite to the intended direction of a biodiversity indicator. T was lower at Chirundu (0.39) than at Walking (0.67) and Cage (0.65), because a natural bird community is silent at night and concentrates its calling around and after dawn (Fig. 12a). The zoo produced steadier detections from resident and captive animals. Multiplying diversity by T removed most of the difference between sites (ABI\textsubscript{rev} 0.150 [0.096–0.214] vs 0.132 [0.096–0.159] and 0.162 [0.129–0.189]).

We therefore recommend reporting T as a separate descriptor of temporal structure, not as part of the diversity index. A more ecologically meaningful stability measure would compare the same hour across days (day-to-day consistency) rather than hours within a day. That requires multi-day series, which our Chirundu data do not provide.

\subsection*{4.4 High-confidence false positives}
Confidence thresholds alone did not remove the largest errors. Tawny Owl detections reached a confidence of 0.999, and 466 of them would have made it one of the most abundant species at the zoo. The location and season meta-model removed them without manual review. In the simulation, false positives at a rate of 30\% raised observed richness from 40 to 100 species, showing how quickly uncontrolled errors inflate detection-based indices.

The sources of these sounds remain to be identified. Most came from the Pool and Walking recorders and could be captive owls, other captive animals or hippopotamus vocalisations. Captive and managed settings, where many non-native animals vocalise, therefore need location filtering and verification of the dominant species, following recommendations for the use of detector scores (Wood and Kahl, 2024). Expert verification of the exported clips will allow species-specific thresholds to replace the global value of 0.50.

Several observations support confidence in the detections that passed both filters. The species that dominate each site are the expected residents: Green-backed Camaroptera, Yellow-bellied Greenbul, Black-backed Puffback, Ring-necked Dove and Arrow-marked Babbler at Chirundu, and Laughing Dove, Eurasian Collared-Dove, House Crow, Eurasian Hoopoe and Common Myna at the zoo, which are the characteristic birds of Abu Dhabi's urban parks. The five most frequent species account for 68\% of the Chirundu detections and 91–93\% of those at Cage and Walking, with mean confidence 0.65–0.83, and the detection-weighted mean confidence is 0.74–0.77 at every site except the Pool. Most retained species were detected in several hours, and the few species that remain unexpected (Section 3.3) have a negligible effect on the index. The remaining uncertainty concerns the precision of rare species, which only expert verification can resolve.

\subsection*{4.5 Evaluation practice for bioacoustic classifiers}
Recording-grouped cross-validation is not optional. A random split of frames inflated accuracy from 0.41 to 0.70, an error large enough to change conclusions about model choice. A classifier trained on Puerto Rican species also failed on African recordings, where only 0.1\% of windows exceeded a confidence of 0.5. Region-specific training sets cannot be transferred to other regions without new labels. Pretrained embeddings from a global model with a simple linear classifier matched or exceeded fine-tuned CNNs at a small fraction of the computing cost, in line with recent work on bioacoustic foundation models (Ghani et al., 2023).

\subsection*{4.6 Limitations}
\begin{enumerate}
\item \textbf{Replication.} Chirundu is represented by one recorder and 6 h over two days, and each zoo habitat by one recorder. Site contrasts therefore rest on temporal, not spatial, replication, and hour-level contrasts with Cage and Walking were not significant.
\item \textbf{Season and year.} The sites were recorded in different seasons and years (wet-season February 2024 vs hot-season July 2025), which confounds habitat with phenology.
\item \textbf{Taxonomic scope.} BirdNET targets birds. Frogs, insects and mammals, which dominate parts of both soundscapes, are not included in the species metrics.
\item \textbf{Verification.} The detections have not yet been verified by an expert, so the precision of most species is unknown.
\item \textbf{Bandwidth.} The sampling rate limits the analysis to 12 kHz, which excludes high-frequency insects and bats.
\item \textbf{Contrast design.} The savanna–zoo comparison is a proof of concept across highly contrasting environments. Gradients of disturbance within natural habitats would be a stronger test of the index's sensitivity.
\end{enumerate}
\subsection*{4.7 Recommendations for practitioners}
\begin{itemize}
\item Evaluate classifiers with folds grouped by recording or site, never by random frames.
\item Apply a location and season filter as well as a confidence threshold, and verify the most frequent species manually.
\item Standardise effort by diel-matched rarefaction before comparing sites.
\item Report \textsuperscript{0}D, \textsuperscript{1}D and \textsuperscript{2}D, ABI\textsubscript{E} with a declared S\textsubscript{ref} and model version, and T separately.
\item Use acoustic indices to describe soundscape structure, not as stand-alone measures of biodiversity.
\end{itemize}
\section*{5. Conclusions}
Species-level detections tracked bird richness and separated a Zambezi Valley savanna from three zoo habitats, which standard acoustic indices did not. Under equal effort the savanna held 21 species and ABI\textsubscript{E} = 0.38, against 1.3–13 species and ABI\textsubscript{E} = 0.03–0.21 at the zoo. A location filter removed high-confidence false positives that a confidence threshold missed, and leakage-free evaluation changed the apparent classifier performance substantially. We recommend the ABI framework of verified detections, a declared regional pool, effort standardisation and Hill numbers, with temporal stability reported as a separate component.

\subsection*{CRediT authorship contribution statement}
\textbf{Muhammad Saeed:} Conceptualisation, Methodology, Software, Formal analysis, Investigation, Validation, Visualisation, Writing – original draft, Writing – review and editing. \textbf{Dmitry Mikhaylov:} Data curation, Software, Supervision, Writing – review and editing. \textbf{Mohamed H. Alhosani:} Supervision, Resources, Writing – review and editing. \textbf{Yasser F. Al Wahedi:} Supervision, Project administration, Writing – review and editing. \textbf{Rory Pilossof:} Investigation (field deployment, Chirundu), Resources, Writing – review and editing. \textbf{Ahmed Arabi:} Resources, Investigation (zoo recordings), Writing – review and editing. \textbf{Walid Shaaba:} Resources, Project administration, Writing – review and editing.

\subsection*{Data and code availability}
The analysis code is released as a Python package (acoustic-biodiversity-index) and will be deposited in a public repository with the journal version; recordings and detection tables are available from the corresponding author on reasonable request. The RFCx data are available from Kaggle (Rainforest Connection Species Audio Detection, 2021). Table S7 (hourly metrics) is available as a CSV file on request.

\subsection*{Declaration of competing interest}
The authors declare that they have no known competing financial interests or personal relationships that could have appeared to influence the work reported in this paper.

\subsection*{Acknowledgements}
We thank My Trees Trust for access to the Chirundu site and for the field deployment of the recorder, and the management and staff of Emirates Park Zoo and Resort for permitting and supporting the recordings.

\section*{References}
Alcocer, I., Lima, H., Sugai, L.S.M., Llusia, D., 2022. Acoustic indices as proxies for biodiversity: a meta-analysis. Biol. Rev. 97, 2209–2236.

Boelman, N.T., Asner, G.P., Hart, P.J., Martin, R.E., 2007. Multi-trophic invasion resistance in Hawaii: bioacoustics, field surveys, and airborne remote sensing. Ecol. Appl. 17, 2137–2144.

Buxton, R.T., McKenna, M.F., Clapp, M., Meyer, E., Stabenau, E., Angeloni, L.M., Crooks, K., Wittemyer, G., 2018. Efficacy of extracting indices from large-scale acoustic recordings to monitor biodiversity. Conserv. Biol. 32, 1174–1184.

Chao, A., Gotelli, N.J., Hsieh, T.C., Sander, E.L., Ma, K.H., Colwell, R.K., Ellison, A.M., 2014. Rarefaction and extrapolation with Hill numbers: a framework for sampling and estimation in species diversity studies. Ecol. Monogr. 84, 45–67.

Cliff, N., 1993. Dominance statistics: ordinal analyses to answer ordinal questions. Psychol. Bull. 114, 494–509.

Fairbrass, A.J., Rennert, P., Williams, C., Titheridge, H., Jones, K.E., 2017. Biases of acoustic indices measuring biodiversity in urban areas. Ecol. Indic. 83, 169–177.

Ghani, B., Denton, T., Kahl, S., Klinck, H., 2023. Global birdsong embeddings enable superior transfer learning for bioacoustic classification. Sci. Rep. 13, 22876.

Gibb, R., Browning, E., Glover-Kapfer, P., Jones, K.E., 2019. Emerging opportunities and challenges for passive acoustics in ecological assessment and monitoring. Methods Ecol. Evol. 10, 169–185.

Hill, M.O., 1973. Diversity and evenness: a unifying notation and its consequences. Ecology 54, 427–432.

Holm, S., 1979. A simple sequentially rejective multiple test procedure. Scand. J. Stat. 6, 65–70.

Jost, L., 2006. Entropy and diversity. Oikos 113, 363–375.

Kaggle, 2021. Rainforest Connection Species Audio Detection. https://www.kaggle.com/c/rfcx-species-audio-detection

Kahl, S., Wood, C.M., Eibl, M., Klinck, H., 2021. BirdNET: a deep learning solution for avian diversity monitoring. Ecol. Inform. 61, 101236.

Kasten, E.P., Gage, S.H., Fox, J., Joo, W., 2012. The remote environmental assessment laboratory's acoustic library: an archive for studying soundscape ecology. Ecol. Inform. 12, 50–67.

Künsch, H.R., 1989. The jackknife and the bootstrap for general stationary observations. Ann. Stat. 17, 1217–1241.

My Trees Trust, 2023. Operations Update 2022/2023. My Trees Trust, Harare, Zimbabwe. https://mytreestrust.org/

Park, D.S., Chan, W., Zhang, Y., Chiu, C.-C., Zoph, B., Cubuk, E.D., Le, Q.V., 2019. SpecAugment: a simple data augmentation method for automatic speech recognition. Proc. Interspeech 2019, 2613–2617.

Pielou, E.C., 1966. The measurement of diversity in different types of biological collections. J. Theor. Biol. 13, 131–144.

Pieretti, N., Farina, A., Morri, D., 2011. A new methodology to infer the singing activity of an avian community: the Acoustic Complexity Index (ACI). Ecol. Indic. 11, 868–873.

Roberts, D.R., Bahn, V., Ciuti, S., Boyce, M.S., Elith, J., Guillera-Arroita, G., Hauenstein, S., Lahoz-Monfort, J.J., Schröder, B., Thuiller, W., Warton, D.I., Wintle, B.A., Hartig, F., Dormann, C.F., 2017. Cross-validation strategies for data with temporal, spatial, hierarchical, or phylogenetic structure. Ecography 40, 913–929.

Sandler, M., Howard, A., Zhu, M., Zhmoginov, A., Chen, L.-C., 2018. MobileNetV2: inverted residuals and linear bottlenecks. Proc. IEEE Conf. Comput. Vis. Pattern Recognit., 4510–4520.

Sethi, S.S., Jones, N.S., Fulcher, B.D., Picinali, L., Clink, D.J., Klinck, H., Orme, C.D.L., Wrege, P.H., Ewers, R.M., 2020. Characterizing soundscapes across diverse ecosystems using a universal acoustic feature set. Proc. Natl. Acad. Sci. U.S.A. 117, 17049–17055.

Stowell, D., 2022. Computational bioacoustics with deep learning: a review and roadmap. PeerJ 10, e13152.

Sueur, J., Pavoine, S., Hamerlynck, O., Duvail, S., 2008. Rapid acoustic survey for biodiversity appraisal. PLoS ONE 3, e4065.

Sugai, L.S.M., Silva, T.S.F., Ribeiro, J.W., Llusia, D., 2019. Terrestrial passive acoustic monitoring: review and perspectives. BioScience 69, 15–25.

Tan, M., Le, Q.V., 2019. EfficientNet: rethinking model scaling for convolutional neural networks. Proc. 36th Int. Conf. Mach. Learn., PMLR 97, 6105–6114.

Towsey, M., Zhang, L., Cottman-Fields, M., Wimmer, J., Zhang, J., Roe, P., 2014. Visualization of long-duration acoustic recordings of the environment. Procedia Comput. Sci. 29, 703–712.

Ulloa, J.S., Haupert, S., Latorre, J.F., Aubin, T., Sueur, J., 2021. scikit-maad: an open-source and modular toolbox for quantitative soundscape analysis in Python. Methods Ecol. Evol. 12, 2334–2340.

Villanueva-Rivera, L.J., Pijanowski, B.C., Doucette, J., Pekin, B., 2011. A primer of acoustic analysis for landscape ecologists. Landsc. Ecol. 26, 1233–1246.

Wood, C.M., Kahl, S., 2024. Guidelines for appropriate use of BirdNET scores and other detector outputs. J. Ornithol. 165, 777–782.

\clearpage
\section*{Appendix A. Supplementary material}
{\small\setlength{\tabcolsep}{4pt}
\begin{longtable}{lc}
\caption{\textbf{Table S1.} Leakage audit: frame-level random-forest accuracy under two evaluation protocols.}\\
\toprule
\textbf{Protocol} & \textbf{Accuracy} \\
\midrule
\endfirsthead
\toprule
\textbf{Protocol} & \textbf{Accuracy} \\
\midrule
\endhead
\bottomrule
\endlastfoot
Frame-level random 80/20 split & 0.703 \\
Frame-level, recording-grouped 5-fold CV & 0.409 $\pm$ 0.022 \\
\end{longtable}
}

{\small\setlength{\tabcolsep}{4pt}
\begin{longtable}{lccc}
\caption{\textbf{Table S2.} Domain shift: maximum softmax probability of the RFCx-trained CNN ensemble in-domain (out-of-fold RFCx) and out-of-domain (Chirundu 10-s windows).}\\
\toprule
\textbf{Data} & \textbf{Windows} & \textbf{Median max. probability} & \textbf{Share > 0.5} \\
\midrule
\endfirsthead
\toprule
\textbf{Data} & \textbf{Windows} & \textbf{Median max. probability} & \textbf{Share > 0.5} \\
\midrule
\endhead
\bottomrule
\endlastfoot
RFCx (in-domain, OOF) & 1216 & 0.82 & 83.4\% \\
Chirundu, Zimbabwe (out-of-domain) & 720 & 0.19 & 0.1\% \\
\end{longtable}
}

{\footnotesize\setlength{\tabcolsep}{4pt}
\begin{longtable}{lcccccc}
\caption{\textbf{Table S3.} Recording inventory and analysed effort.}\\
\toprule
\textbf{Site} & \textbf{Files} & \textbf{First file} & \textbf{Last file} & \textbf{Sample rate (Hz)} & \textbf{Recorded (h)} & \textbf{Analysed (h)} \\
\midrule
\endfirsthead
\toprule
\textbf{Site} & \textbf{Files} & \textbf{First file} & \textbf{Last file} & \textbf{Sample rate (Hz)} & \textbf{Recorded (h)} & \textbf{Analysed (h)} \\
\midrule
\endhead
\bottomrule
\endlastfoot
Chirundu & 12 & 2024-02-01 15:11:02 & 2024-02-02 11:11:02 & 24000 & 12.0 & 6.0 \\
Zoo – Walking & 55 & 2025-07-27 02:07:29 & 2025-07-29 08:07:11 & 24000 & 54.0 & 27.0 \\
Zoo – Cage & 27 & 2025-07-27 02:07:32 & 2025-07-28 04:07:30 & 24000 & 26.1 & 13.1 \\
Zoo – Pool & 54 & 2025-07-27 02:52:36 & 2025-07-29 07:52:40 & 24000 & 53.3 & 26.8 \\
\end{longtable}
}

{\footnotesize\setlength{\tabcolsep}{4pt}
\begin{longtable}{lcccccccc}
\caption{\textbf{Table S4.} Sensitivity to the BirdNET confidence threshold (selected thresholds; full table in the data deposit).}\\
\toprule
\textbf{Site} & \textbf{Threshold} & \textbf{S} & \textbf{\textsuperscript{1}D} & \textbf{J$'$} & \textbf{ABI\textsubscript{E}} & \textbf{T} & \textbf{ABI\textsubscript{rev}} & \textbf{ABI\textsubscript{orig}} \\
\midrule
\endfirsthead
\toprule
\textbf{Site} & \textbf{Threshold} & \textbf{S} & \textbf{\textsuperscript{1}D} & \textbf{J$'$} & \textbf{ABI\textsubscript{E}} & \textbf{T} & \textbf{ABI\textsubscript{rev}} & \textbf{ABI\textsubscript{orig}} \\
\midrule
\endhead
\bottomrule
\endlastfoot
Chirundu & 0.10 & 62 & 18.65 & 0.71 & 0.49 & 0.47 & 0.228 & 0.195 \\
Chirundu & 0.30 & 31 & 12.16 & 0.73 & 0.42 & 0.43 & 0.178 & 0.198 \\
Chirundu & 0.50 & 21 & 9.83 & 0.75 & 0.38 & 0.39 & 0.150 & 0.205 \\
Chirundu & 0.70 & 15 & 8.96 & 0.81 & 0.36 & 0.32 & 0.117 & 0.215 \\
Chirundu & 0.90 & 10 & 6.86 & 0.84 & 0.32 & 0.25 & 0.080 & 0.220 \\
Zoo – Walking & 0.10 & 89 & 13.86 & 0.59 & 0.44 & 0.67 & 0.293 & 0.094 \\
Zoo – Walking & 0.30 & 40 & 6.42 & 0.50 & 0.31 & 0.67 & 0.206 & 0.077 \\
Zoo – Walking & 0.50 & 25 & 4.26 & 0.45 & 0.24 & 0.67 & 0.162 & 0.069 \\
Zoo – Walking & 0.70 & 16 & 3.34 & 0.43 & 0.20 & 0.68 & 0.136 & 0.070 \\
Zoo – Walking & 0.90 & 8 & 2.41 & 0.42 & 0.15 & 0.63 & 0.092 & 0.062 \\
Zoo – Cage & 0.10 & 46 & 8.02 & 0.54 & 0.35 & 0.64 & 0.221 & 0.112 \\
Zoo – Cage & 0.30 & 24 & 4.72 & 0.49 & 0.26 & 0.64 & 0.166 & 0.097 \\
Zoo – Cage & 0.50 & 17 & 3.37 & 0.43 & 0.20 & 0.65 & 0.132 & 0.088 \\
Zoo – Cage & 0.70 & 10 & 2.59 & 0.41 & 0.16 & 0.66 & 0.104 & 0.088 \\
Zoo – Cage & 0.90 & 7 & 2.53 & 0.48 & 0.15 & 0.61 & 0.094 & 0.101 \\
Zoo – Pool & 0.10 & 36 & 5.30 & 0.47 & 0.28 & 0.69 & 0.192 & 0.076 \\
Zoo – Pool & 0.30 & 13 & 5.40 & 0.66 & 0.28 & 0.53 & 0.148 & 0.094 \\
Zoo – Pool & 0.50 & 4 & 2.73 & 0.72 & 0.17 & 0.36 & 0.060 & 0.098 \\
Zoo – Pool & 0.70 & 2 & 2.00 & 1.00 & 0.12 & $-$0.00 & -0.000 & 0.120 \\
Zoo – Pool & 0.90 & 0 & nan & nan & nan & nan & nan & nan \\
\end{longtable}
}

{\footnotesize\setlength{\tabcolsep}{4pt}
\begin{longtable}{lcccccc}
\caption{\textbf{Table S5.} Sensitivity to the time-bin width.}\\
\toprule
\textbf{Site} & \textbf{Bin (min)} & \textbf{Bins} & \textbf{ABI\textsubscript{E}} & \textbf{T} & \textbf{ABI\textsubscript{rev}} & \textbf{ABI\textsubscript{orig}} \\
\midrule
\endfirsthead
\toprule
\textbf{Site} & \textbf{Bin (min)} & \textbf{Bins} & \textbf{ABI\textsubscript{E}} & \textbf{T} & \textbf{ABI\textsubscript{rev}} & \textbf{ABI\textsubscript{orig}} \\
\midrule
\endhead
\bottomrule
\endlastfoot
Chirundu & 15 & 36 & 0.38 & 0.45 & 0.170 & 0.137 \\
Chirundu & 30 & 24 & 0.38 & 0.44 & 0.169 & 0.161 \\
Chirundu & 60 & 12 & 0.38 & 0.39 & 0.150 & 0.205 \\
Chirundu & 120 & 7 & 0.38 & 0.36 & 0.135 & 0.251 \\
Zoo – Walking & 15 & 163 & 0.24 & 0.71 & 0.170 & 0.044 \\
Zoo – Walking & 30 & 109 & 0.24 & 0.69 & 0.166 & 0.052 \\
Zoo – Walking & 60 & 55 & 0.24 & 0.67 & 0.162 & 0.069 \\
Zoo – Walking & 120 & 28 & 0.24 & 0.63 & 0.152 & 0.091 \\
Zoo – Cage & 15 & 79 & 0.20 & 0.65 & 0.132 & 0.056 \\
Zoo – Cage & 30 & 53 & 0.20 & 0.60 & 0.122 & 0.064 \\
Zoo – Cage & 60 & 27 & 0.20 & 0.65 & 0.132 & 0.088 \\
Zoo – Cage & 120 & 14 & 0.20 & 0.61 & 0.122 & 0.113 \\
Zoo – Pool & 15 & 161 & 0.17 & 0.38 & 0.063 & 0.061 \\
Zoo – Pool & 30 & 108 & 0.17 & 0.36 & 0.060 & 0.073 \\
Zoo – Pool & 60 & 55 & 0.17 & 0.36 & 0.060 & 0.098 \\
Zoo – Pool & 120 & 28 & 0.17 & 0.37 & 0.062 & 0.132 \\
\end{longtable}
}

{\footnotesize\setlength{\tabcolsep}{4pt}
\begin{longtable}{lccccccc}
\caption{\textbf{Table S6.} Simulation results (means of 40 replicates).}\\
\toprule
\textbf{Scenario} & \textbf{Value} & \textbf{Observed S} & \textbf{J$'$} & \textbf{ABI\textsubscript{E}} & \textbf{T} & \textbf{ABI\textsubscript{rev}} & \textbf{ABI\textsubscript{orig}} \\
\midrule
\endfirsthead
\toprule
\textbf{Scenario} & \textbf{Value} & \textbf{Observed S} & \textbf{J$'$} & \textbf{ABI\textsubscript{E}} & \textbf{T} & \textbf{ABI\textsubscript{rev}} & \textbf{ABI\textsubscript{orig}} \\
\midrule
\endhead
\bottomrule
\endlastfoot
Richness S & 5 & 5.0 & 0.81 & 0.28 & 0.89 & 0.25 & 0.52 \\
Richness S & 10 & 10.0 & 0.80 & 0.40 & 0.89 & 0.36 & 0.49 \\
Richness S & 20 & 20.0 & 0.86 & 0.56 & 0.87 & 0.49 & 0.49 \\
Richness S & 40 & 40.0 & 0.87 & 0.70 & 0.85 & 0.59 & 0.44 \\
Richness S & 60 & 59.9 & 0.89 & 0.79 & 0.84 & 0.66 & 0.42 \\
Richness S & 80 & 79.7 & 0.88 & 0.84 & 0.82 & 0.69 & 0.39 \\
Richness S & 100 & 99.5 & 0.89 & 0.89 & 0.81 & 0.72 & 0.37 \\
Unevenness $\sigma$ & 0 & 40.0 & 1.00 & 0.80 & 0.85 & 0.68 & 0.58 \\
Unevenness $\sigma$ & 0.5 & 40.0 & 0.96 & 0.77 & 0.85 & 0.66 & 0.54 \\
Unevenness $\sigma$ & 1 & 40.0 & 0.87 & 0.70 & 0.85 & 0.60 & 0.45 \\
Unevenness $\sigma$ & 1.5 & 39.4 & 0.74 & 0.59 & 0.86 & 0.51 & 0.32 \\
Unevenness $\sigma$ & 2 & 36.8 & 0.64 & 0.50 & 0.86 & 0.43 & 0.24 \\
Unevenness $\sigma$ & 2.5 & 32.1 & 0.57 & 0.43 & 0.87 & 0.37 & 0.20 \\
Unevenness $\sigma$ & 3 & 25.9 & 0.50 & 0.36 & 0.88 & 0.32 & 0.17 \\
Temporal CV & 0.1 & 40.0 & 0.87 & 0.70 & 0.90 & 0.63 & 0.48 \\
Temporal CV & 0.25 & 40.0 & 0.87 & 0.70 & 0.89 & 0.62 & 0.47 \\
Temporal CV & 0.5 & 40.0 & 0.88 & 0.70 & 0.85 & 0.60 & 0.45 \\
Temporal CV & 0.75 & 40.0 & 0.87 & 0.70 & 0.82 & 0.57 & 0.42 \\
Temporal CV & 1 & 40.0 & 0.87 & 0.69 & 0.78 & 0.54 & 0.38 \\
Temporal CV & 1.5 & 40.0 & 0.86 & 0.69 & 0.70 & 0.48 & 0.33 \\
False-positive rate & 0 & 40.0 & 0.86 & 0.69 & 0.85 & 0.59 & 0.44 \\
False-positive rate & 0.02 & 66.9 & 0.79 & 0.72 & 0.85 & 0.61 & 0.24 \\
False-positive rate & 0.05 & 85.3 & 0.76 & 0.73 & 0.84 & 0.61 & 0.22 \\
False-positive rate & 0.1 & 97.0 & 0.76 & 0.75 & 0.84 & 0.63 & 0.24 \\
False-positive rate & 0.2 & 99.8 & 0.80 & 0.80 & 0.83 & 0.67 & 0.31 \\
False-positive rate & 0.3 & 100.0 & 0.83 & 0.83 & 0.84 & 0.69 & 0.36 \\
\end{longtable}
}

{\small\setlength{\tabcolsep}{4pt}
\begin{longtable}{p{2.3cm}p{6.4cm}ccc}
\caption{\textbf{Table S8.} Species removed by the strict location filter (meta-model score < 0.10).}\\
\toprule
\textbf{Site} & \textbf{Species} & \textbf{Detections} & \textbf{Max. confidence} & \textbf{Location score} \\
\midrule
\endfirsthead
\toprule
\textbf{Site} & \textbf{Species} & \textbf{Detections} & \textbf{Max. confidence} & \textbf{Location score} \\
\midrule
\endhead
\bottomrule
\endlastfoot
Zoo – Cage & Tawny Owl (\textit{Strix aluco}) & 24 & 0.87 & 0.03 \\
Zoo – Cage & Rock Sparrow (\textit{Petronia petronia}) & 1 & 0.51 & 0.03 \\
Zoo – Pool & Tawny Owl (\textit{Strix aluco}) & 135 & 0.99 & 0.03 \\
Zoo – Pool & Great Spotted Woodpecker (\textit{Dendrocopos major}) & 41 & 0.95 & 0.03 \\
Zoo – Pool & Redwing (\textit{Turdus iliacus}) & 2 & 0.76 & 0.03 \\
Zoo – Pool & Common Buzzard (\textit{Buteo buteo}) & 1 & 0.67 & 0.05 \\
Zoo – Walking & Tawny Owl (\textit{Strix aluco}) & 307 & 1.00 & 0.03 \\
Zoo – Walking & Common Buzzard (\textit{Buteo buteo}) & 1 & 0.94 & 0.05 \\
Zoo – Walking & Great Spotted Woodpecker (\textit{Dendrocopos major}) & 1 & 0.93 & 0.03 \\
Zoo – Walking & Reed Bunting (\textit{Emberiza schoeniclus}) & 1 & 0.52 & 0.03 \\
Zoo – Walking & Ring-necked Pheasant (\textit{Phasianus colchicus}) & 1 & 0.57 & 0.05 \\
Zoo – Walking & European Golden-Plover (\textit{Pluvialis apricaria}) & 1 & 0.92 & 0.05 \\
Zoo – Walking & Eurasian Bullfinch (\textit{Pyrrhula pyrrhula}) & 1 & 0.58 & 0.03 \\
\end{longtable}
}

{\footnotesize\setlength{\tabcolsep}{4pt}
\begin{longtable}{lcccccc}
\caption{\textbf{Table S9.} Effort-standardised, diel-matched comparison (reference: Chirundu, 6.0 h). Zoo values are means of 500 draws with 2.5–97.5\% ranges; Chirundu shows block-bootstrap 95\% CIs.}\\
\toprule
\textbf{Site} & \textbf{Effort (h)} & \textbf{S} & \textbf{\textsuperscript{1}D} & \textbf{ABI\textsubscript{E}} & \textbf{T} & \textbf{ABI\textsubscript{rev}} \\
\midrule
\endfirsthead
\toprule
\textbf{Site} & \textbf{Effort (h)} & \textbf{S} & \textbf{\textsuperscript{1}D} & \textbf{ABI\textsubscript{E}} & \textbf{T} & \textbf{ABI\textsubscript{rev}} \\
\midrule
\endhead
\bottomrule
\endlastfoot
Chirundu & 6.0 & 21.0 [11.0–21.0] & 9.83 [4.48–11.90] & 0.38 [0.25–0.41] & 0.39 [0.29–0.61] & 0.150 [0.096–0.214] \\
Zoo – Walking & 5.8 & 12.6 [10.0–15.0] & 3.51 [2.68–4.62] & 0.21 [0.16–0.25] & 0.54 [0.42–0.62] & 0.110 [0.096–0.125] \\
Zoo – Cage & 5.8 & 13.0 & 2.93 & 0.18 & 0.60 & 0.108 \\
Zoo – Pool & 5.8 & 1.3 [1.0–2.0] & 1.28 [1.00–2.00] & 0.03 [0.00–0.12] & 0.11 [0.00–0.48] & 0.001 [0.000–0.023] \\
\end{longtable}
}

{\footnotesize\setlength{\tabcolsep}{4pt}
\begin{longtable}{lcccccc}
\caption{\textbf{Table S10.} Sensitivity to the location-filter threshold (columns): richness S / ABI\textsubscript{E} per site, and the regional pool S\textsubscript{ref} at each threshold.}\\
\toprule
\textbf{Site} & \textbf{0.03} & \textbf{0.05} & \textbf{0.10} & \textbf{0.15} & \textbf{0.20} & \textbf{0.30} \\
\midrule
\endfirsthead
\toprule
\textbf{Site} & \textbf{0.03} & \textbf{0.05} & \textbf{0.10} & \textbf{0.15} & \textbf{0.20} & \textbf{0.30} \\
\midrule
\endhead
\bottomrule
\endlastfoot
Chirundu & 21 / 0.37 & 21 / 0.38 & 21 / 0.38 & 21 / 0.38 & 21 / 0.38 & 21 / 0.39 \\
Zoo – Walking & 32 / 0.27 & 28 / 0.24 & 25 / 0.24 & 24 / 0.24 & 22 / 0.24 & 20 / 0.21 \\
Zoo – Cage & 19 / 0.22 & 17 / 0.20 & 17 / 0.20 & 15 / 0.20 & 14 / 0.20 & 11 / 0.18 \\
Zoo – Pool & 8 / 0.14 & 5 / 0.20 & 4 / 0.17 & 4 / 0.17 & 4 / 0.17 & 4 / 0.17 \\
S\textsubscript{ref} & 455 & 432 & 409 & 394 & 383 & 359 \\
\end{longtable}
}

{\footnotesize\setlength{\tabcolsep}{4pt}
\begin{longtable}{p{2.0cm}p{6.0cm}cccc}
\caption{\textbf{Table S11.} All species detected per site (confidence $\geq$ 0.50, strict location filter).}\\
\toprule
\textbf{Site} & \textbf{Species} & \textbf{Detections} & \textbf{Max. conf.} & \textbf{Mean conf.} & \textbf{Hours present} \\
\midrule
\endfirsthead
\toprule
\textbf{Site} & \textbf{Species} & \textbf{Detections} & \textbf{Max. conf.} & \textbf{Mean conf.} & \textbf{Hours present} \\
\midrule
\endhead
\bottomrule
\endlastfoot
Chirundu & Green-backed Camaroptera (\textit{Camaroptera brachyura}) & 94 & 0.99 & 0.74 & 5 \\
Chirundu & Yellow-bellied Greenbul (\textit{Chlorocichla flaviventris}) & 29 & 0.98 & 0.73 & 3 \\
Chirundu & Black-backed Puffback (\textit{Dryoscopus cubla}) & 25 & 0.99 & 0.73 & 6 \\
Chirundu & Ring-necked Dove (\textit{Streptopelia capicola}) & 24 & 0.97 & 0.69 & 7 \\
Chirundu & Arrow-marked Babbler (\textit{Turdoides jardineii}) & 23 & 0.99 & 0.81 & 1 \\
Chirundu & Eastern Nicator (\textit{Nicator gularis}) & 21 & 0.94 & 0.66 & 3 \\
Chirundu & African Fish-Eagle (\textit{Haliaeetus vocifer}) & 19 & 1.00 & 0.84 & 5 \\
Chirundu & Green Woodhoopoe (\textit{Phoeniculus purpureus}) & 18 & 1.00 & 0.89 & 2 \\
Chirundu & Brown-hooded Kingfisher (\textit{Halcyon albiventris}) & 9 & 0.91 & 0.79 & 1 \\
Chirundu & Meves's Starling (\textit{Lamprotornis mevesii}) & 5 & 0.73 & 0.59 & 3 \\
Chirundu & White-browed Sparrow-Weaver (\textit{Plocepasser mahali}) & 4 & 0.94 & 0.78 & 1 \\
Chirundu & African Gray Hornbill (\textit{Lophoceros nasutus}) & 3 & 0.74 & 0.63 & 1 \\
Chirundu & Terrestrial Brownbul (\textit{Phyllastrephus terrestris}) & 3 & 0.82 & 0.76 & 2 \\
Chirundu & Fork-tailed Drongo (\textit{Dicrurus adsimilis}) & 2 & 0.53 & 0.53 & 2 \\
Chirundu & European Bee-eater (\textit{Merops apiaster}) & 2 & 0.63 & 0.60 & 1 \\
Chirundu & African Goshawk (\textit{Accipiter tachiro}) & 1 & 0.88 & 0.88 & 1 \\
Chirundu & Bearded Scrub-Robin (\textit{Cercotrichas quadrivirgata}) & 1 & 0.54 & 0.54 & 1 \\
Chirundu & White-browed Robin-Chat (\textit{Cossypha heuglini}) & 1 & 0.58 & 0.58 & 1 \\
Chirundu & Crested Barbet (\textit{Trachyphonus vaillantii}) & 1 & 0.70 & 0.70 & 1 \\
Chirundu & Emerald-spotted Wood-Dove (\textit{Turtur chalcospilos}) & 1 & 0.53 & 0.53 & 1 \\
Chirundu & Eurasian Hoopoe (\textit{Upupa epops}) & 1 & 0.76 & 0.76 & 1 \\
Zoo – Walking & Laughing Dove (\textit{Streptopelia senegalensis}) & 431 & 0.99 & 0.79 & 13 \\
Zoo – Walking & Eurasian Collared-Dove (\textit{Streptopelia decaocto}) & 73 & 1.00 & 0.82 & 10 \\
Zoo – Walking & House Crow (\textit{Corvus splendens}) & 53 & 0.97 & 0.69 & 6 \\
Zoo – Walking & Common Cuckoo (\textit{Cuculus canorus}) & 38 & 0.90 & 0.69 & 9 \\
Zoo – Walking & Common Myna (\textit{Acridotheres tristis}) & 22 & 0.92 & 0.69 & 3 \\
Zoo – Walking & Red-wattled Lapwing (\textit{Vanellus indicus}) & 12 & 1.00 & 0.84 & 4 \\
Zoo – Walking & Mallard (\textit{Anas platyrhynchos}) & 11 & 0.89 & 0.61 & 5 \\
Zoo – Walking & Barn Owl (\textit{Tyto alba}) & 8 & 0.98 & 0.78 & 1 \\
Zoo – Walking & Eurasian Hoopoe (\textit{Upupa epops}) & 4 & 0.72 & 0.70 & 4 \\
Zoo – Walking & Great Bittern (\textit{Botaurus stellaris}) & 3 & 0.73 & 0.68 & 3 \\
Zoo – Walking & Purple Sunbird (\textit{Cinnyris asiaticus}) & 3 & 0.75 & 0.65 & 1 \\
Zoo – Walking & Eurasian Moorhen (\textit{Gallinula chloropus}) & 3 & 0.88 & 0.72 & 2 \\
Zoo – Walking & Eurasian Oystercatcher (\textit{Haematopus ostralegus}) & 2 & 0.57 & 0.56 & 1 \\
Zoo – Walking & Eurasian Curlew (\textit{Numenius arquata}) & 2 & 0.72 & 0.68 & 1 \\
Zoo – Walking & Gray Francolin (\textit{Ortygornis pondicerianus}) & 2 & 0.91 & 0.80 & 1 \\
Zoo – Walking & Spotted Crake (\textit{Porzana porzana}) & 2 & 0.69 & 0.63 & 2 \\
Zoo – Walking & Eurasian Nightjar (\textit{Caprimulgus europaeus}) & 1 & 0.55 & 0.55 & 1 \\
Zoo – Walking & Eurasian Dotterel (\textit{Charadrius morinellus}) & 1 & 0.62 & 0.62 & 1 \\
Zoo – Walking & Peregrine Falcon (\textit{Falco peregrinus}) & 1 & 0.79 & 0.79 & 1 \\
Zoo – Walking & Eurasian Coot (\textit{Fulica atra}) & 1 & 0.69 & 0.69 & 1 \\
Zoo – Walking & Whimbrel (\textit{Numenius phaeopus}) & 1 & 0.87 & 0.87 & 1 \\
Zoo – Walking & Black-bellied Plover (\textit{Pluvialis squatarola}) & 1 & 0.55 & 0.55 & 1 \\
Zoo – Walking & Red-vented Bulbul (\textit{Pycnonotus cafer}) & 1 & 0.53 & 0.53 & 1 \\
Zoo – Walking & White-eared Bulbul (\textit{Pycnonotus leucotis}) & 1 & 0.68 & 0.68 & 1 \\
Zoo – Walking & Song Thrush (\textit{Turdus philomelos}) & 1 & 0.59 & 0.59 & 1 \\
Zoo – Cage & Laughing Dove (\textit{Streptopelia senegalensis}) & 317 & 0.99 & 0.77 & 14 \\
Zoo – Cage & Eurasian Collared-Dove (\textit{Streptopelia decaocto}) & 60 & 1.00 & 0.83 & 9 \\
Zoo – Cage & Eurasian Hoopoe (\textit{Upupa epops}) & 24 & 0.94 & 0.70 & 6 \\
Zoo – Cage & House Crow (\textit{Corvus splendens}) & 13 & 0.95 & 0.68 & 3 \\
Zoo – Cage & Mallard (\textit{Anas platyrhynchos}) & 11 & 0.88 & 0.65 & 7 \\
Zoo – Cage & Common Cuckoo (\textit{Cuculus canorus}) & 10 & 0.94 & 0.61 & 6 \\
Zoo – Cage & Common Myna (\textit{Acridotheres tristis}) & 7 & 0.96 & 0.83 & 2 \\
Zoo – Cage & Purple Sunbird (\textit{Cinnyris asiaticus}) & 4 & 0.61 & 0.56 & 3 \\
Zoo – Cage & White-eared Bulbul (\textit{Pycnonotus leucotis}) & 3 & 0.73 & 0.68 & 1 \\
Zoo – Cage & Red-wattled Lapwing (\textit{Vanellus indicus}) & 3 & 0.93 & 0.90 & 2 \\
Zoo – Cage & Graylag Goose (\textit{Anser anser}) & 1 & 0.52 & 0.52 & 1 \\
Zoo – Cage & Tree Pipit (\textit{Anthus trivialis}) & 1 & 0.74 & 0.74 & 1 \\
Zoo – Cage & Rock Pigeon (\textit{Columba livia}) & 1 & 0.53 & 0.53 & 1 \\
Zoo – Cage & Common Wood-Pigeon (\textit{Columba palumbus}) & 1 & 0.58 & 0.58 & 1 \\
Zoo – Cage & Common House-Martin (\textit{Delichon urbicum}) & 1 & 0.62 & 0.62 & 1 \\
Zoo – Cage & Eurasian Wigeon (\textit{Mareca penelope}) & 1 & 0.67 & 0.67 & 1 \\
Zoo – Cage & Eurasian Blackbird (\textit{Turdus merula}) & 1 & 0.57 & 0.57 & 1 \\
Zoo – Pool & Mallard (\textit{Anas platyrhynchos}) & 6 & 0.79 & 0.61 & 5 \\
Zoo – Pool & Dunlin (\textit{Calidris alpina}) & 1 & 0.86 & 0.86 & 1 \\
Zoo – Pool & Eurasian Collared-Dove (\textit{Streptopelia decaocto}) & 1 & 0.62 & 0.62 & 1 \\
Zoo – Pool & Barn Owl (\textit{Tyto alba}) & 1 & 0.61 & 0.61 & 1 \\
\end{longtable}
}

\begin{figure}[!htbp]\centering
\includegraphics[width=\textwidth,height=0.78\textheight,keepaspectratio]{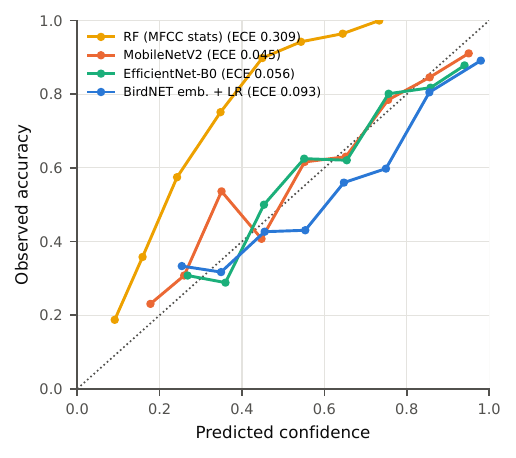}\par\medskip
\caption{\textbf{Fig. S1.} Reliability diagrams of the four classifiers (out-of-fold predictions; ECE in the legend).}
\end{figure}

\begin{figure}[!htbp]\centering
\includegraphics[width=\textwidth,height=0.78\textheight,keepaspectratio]{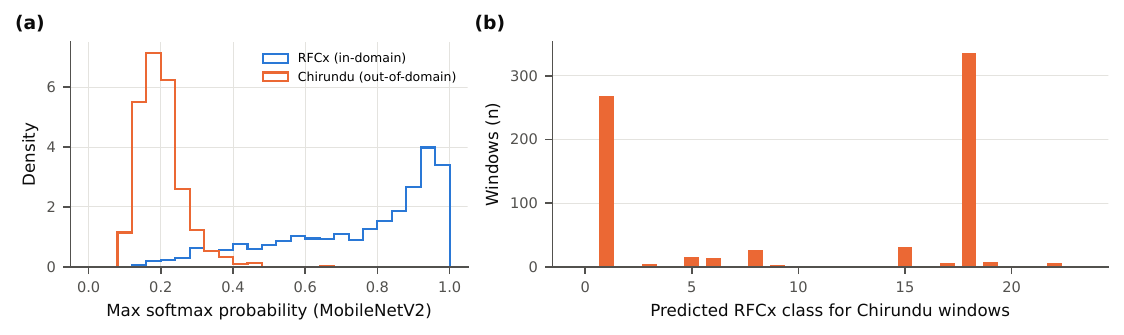}\par\medskip
\caption{\textbf{Fig. S2.} Domain shift. (a) Distribution of the maximum softmax probability of the RFCx-trained CNN ensemble on RFCx (in-domain) and Chirundu (out-of-domain) windows. (b) RFCx classes predicted for Chirundu windows.}
\end{figure}

\begin{figure}[!htbp]\centering
\includegraphics[width=\textwidth,height=0.78\textheight,keepaspectratio]{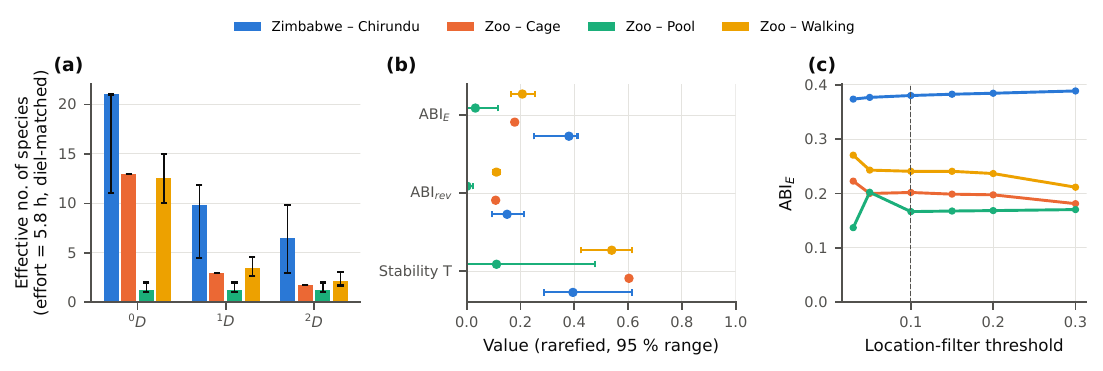}\par\medskip
\caption{\textbf{Fig. S3.} Robustness. (a) Hill numbers and (b) T, ABI\textsubscript{rev} and ABI\textsubscript{E} after diel-matched rarefaction to Chirundu's effort (mean and 2.5–97.5\% range of 500 draws). (c) ABI\textsubscript{E} against the location-filter threshold; the dashed line marks the threshold used (0.10).}
\end{figure}

\end{document}